\documentclass[nohyperref]{article}

\usepackage{microtype}
\usepackage{graphicx}
\usepackage{subfigure}
\usepackage{booktabs} 

\usepackage{hyperref}

\usepackage[accepted]{icml2022}

\usepackage{amsmath}
\usepackage{amssymb}
\usepackage{mathtools}
\usepackage{amsthm}

\usepackage[capitalize,noabbrev]{cleveref}

\theoremstyle{plain}

\theoremstyle{definition}

\theoremstyle{remark}

\usepackage[textsize=tiny]{todonotes}

\icmltitlerunning{On the Effects of Artificial Data Modification}

\usepackage{algorithm}
\usepackage{algorithmic}

\usepackage{tabularx}
\usepackage{tabulary}
\usepackage{multirow}
\newcolumntype{Y}{>{\centering\arraybackslash}X}
\usepackage{xcolor}
\definecolor{blue}{HTML}{377EB8}
\definecolor{green}{HTML}{4DAF4A}

\usepackage{amssymb}

\usepackage{bm}
\usepackage{chngcntr}
\usepackage{float}
\usepackage{stfloats} 

\newcommand{\di}{$DI$\xspace}
\newcommand{\av}[2][\,]{\mathbb{E}_{#1}\!\left[ {\strut #2} \right]}
\usepackage{xspace}
\newcommand{\fmix}{FMix\xspace}

\newcommand{\iocc}{iOcclusion\xspace}
\newcommand{\cutocc}{CutOcclusion\xspace}
\newcommand{\mixup}{MixUp\xspace}
\newcommand{\cutmix}{CutMix\xspace}
\newcommand{\cutout}{CutOut\xspace}

\newcommand{\cifar}[1]{CIFAR-{#1}\xspace}

\newcommand{\imagenet}{ImageNet\xspace}

\makeatletter
\newcommand*{\addFileDependency}[1]{
  \typeout{(#1)}
  \@addtofilelist{#1}
  \IfFileExists{#1}{}{\typeout{No file #1.}}
}
\makeatother

\begin{document}

\twocolumn[
\icmltitle{On the Effects of Artificial Data Modification}




\begin{icmlauthorlist}
\icmlauthor{Antonia Marcu}{yyy}
\icmlauthor{Adam Pr\"ugel-Bennett}{yyy}
\end{icmlauthorlist}

\icmlaffiliation{yyy}{Vision, Learning and Control research group, University of Southampton}

\icmlcorrespondingauthor{Antonia Marcu}{am1g15@soton.ac.uk}


\vskip 0.3in
]



\printAffiliationsAndNotice{} 

\begin{abstract}
    Data distortion is commonly applied in vision models during both training (e.g methods like MixUp and CutMix) and evaluation (e.g. shape-texture bias and robustness).
    This data modification can introduce artificial information.
    It is often assumed that the resulting artefacts are detrimental to training, whilst being negligible when analysing models.
    We investigate these assumptions and conclude that in some cases they are unfounded and lead to incorrect results.
    Specifically, we show current shape bias identification methods and occlusion robustness measures are biased and propose a fairer alternative for the latter. 
    Subsequently, through a series of experiments we seek to correct and strengthen the community's perception of how augmenting affects learning of vision models. Based on our empirical results we argue that the impact of the artefacts must be understood and exploited rather than eliminated.
\end{abstract}

\section{Motivation}
\label{sec:motivation}

Augmentation is commonplace when training models.
It is a form of data modification where samples are artificially distorted to create larger training sets.
Apart from augmentative purposes, data modification is also used for a wide range of model analysis methods. 
Most recently, distortion-based approaches have been adopted when trying to answer key machine learning questions. 
To this end, \mixup-like distortions~\cite{zhang2018mixup} were proposed for empirically predicting generalisation~\cite{schiff2021gi, natekar2020representation, lassance2020ranking}.
Thus, data modification is becoming increasingly popular, but little attention is paid to the secondary effects of this practice.
As we will demonstrate, \textit{our current understanding of the effects of data modification lies on fundamentally flawed assumptions}.
This impacts not only our perception of what features are important to our models, but also the correctness of the distortion-based approaches we propose as a field.

\begin{figure}
\vskip 0.2in
\centering

    \includegraphics[width=0.99\linewidth]{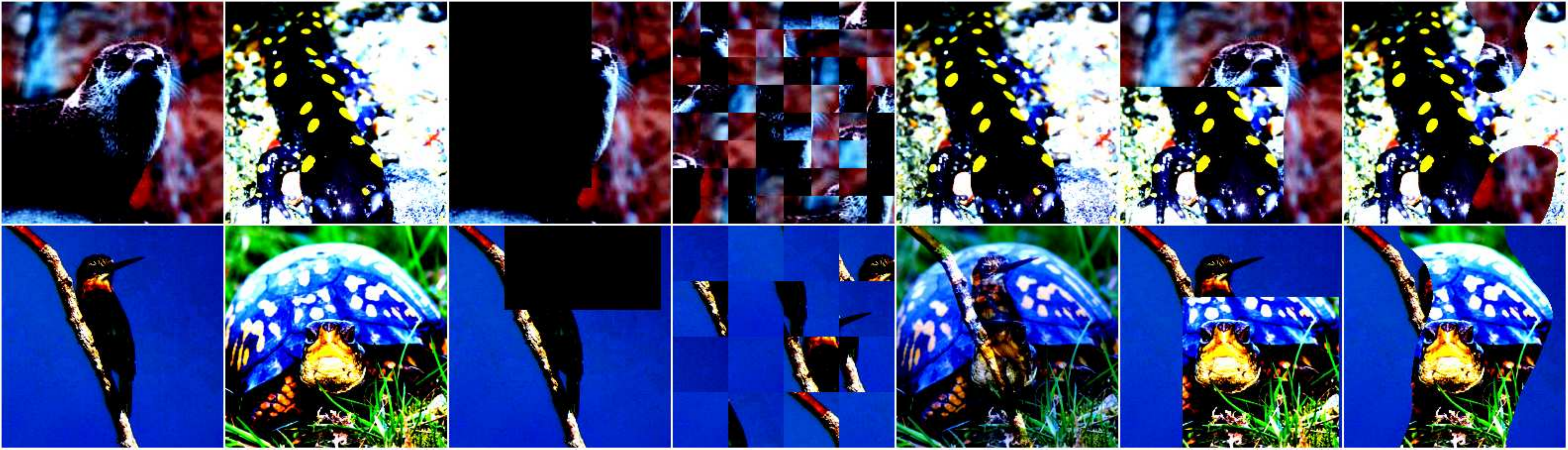}
    \caption{Examples of image distortions. Left to right: source image 1, source image 2, patch occlusion, shuffled grid, \mixup, \cutmix, \fmix.
    For mixing augmentations, the first row was generated with a mixing factor of 0.2, while the second one with 0.5.
    For patch-shuffling, a grid size of $8 \times 8$ was used for the top row and $4 \times 4$ for the bottom one.
    }
    \label{fig:distortions}
    \vskip -0.2in
\end{figure}

In this paper we study the assumptions and implications that arise through artificially distorted data. 
From the model analysis perspective, we take occlusion robustness and shape bias identification methods as examples of where modified data is used. On the training side, we focus on some instances of Mixed Sample Data Augmentation (MSDA), where two images are combined to obtain a new training sample.
Visual illustrations of each can be found in~\cref{fig:distortions}.
We explore some of the \textit{side-effects} of data modification and demonstrate that this practice has resulted in the creation of biased model interpretation tools and poorly informed theories.
More specifically, we study a number of assumptions which we show are erroneous and which lie at the heart of the methods we briefly introduce below. Contesting these assumptions has broader implications on the community's perception of what aspects of the data are important and how distortions can be further used to understand generalisation.

\textbf{Shape-texture bias:}
Deep models are known to be sensitive to distribution shifts \cite{alaiz2008assessing, cieslak2009framework, engstrom2019exploring} and interventions that are imperceptible to humans \cite{szegedy2013intriguing,goodfellow2014explaining}.
It has been argued that this is intimately linked to networks tending to use texture rather than shape information \cite{brendel2018approximating, geirhos2018imagenet}.
Recently, input distortions have become a popular way of assessing a model's texture bias.
To this end, images are divided into a grid and the resulting patches are randomly shuffled such that information is preserved locally, while the global shape is altered \cite{shi2020informative, mummadi2021does, luo2019defective, zhang2019interpreting}.
\textit{It is implicitly assumed that patch-shuffling does not introduce misleading shape or texture that could affect model evaluation.} As such, if a model's accuracy drops when evaluated on patch-shuffled images, this degradation in performance is entirely attributed to the model's bias for shape information. Thus, any side-effects of the data manipulation process are considered negligible.

\textbf{Occlusion robustness:}
Commonly, occlusion robustness is concerned with the amount of information that can be hidden from a model without affecting its ability to classify \cite{tang2018recurrent, rajaei2019beyond}. 
A widely adopted proxy for measuring occlusion robustness is through the raw accuracy obtained after superimposing a rectangular patch on an image \cite{chun2020empirical, BMVC2016_137, yun2019cutmix, zhong2020random, kokhlikyan2020captum}. 
We refer to this approach as \cutocc.
Just as with shape bias this method relies on the information introduced not to interfere with a model's learnt representations, such that a decrease in performance can be directly attributed to a lack of robustness. 
Thus, using \cutocc, one implicitly assumes that artefacts do not interfere with the results of robustness evaluation.  

\textbf{Data augmentation studies:}
In statistical learning, training with augmented data is termed Vicinal Risk Minimisation (VRM) \cite{vapnik2013nature, chapelle2001vicinal} and is seen as injecting prior knowledge about the neighbourhood of the data samples.
The intuition behind augmentation caused researchers to interpret its effect through the similarity between original and augmented data distributions.
This perspective is often challenged by methods which despite generating samples that do not appear to fall under the distribution of natural images, lead to strong learners.
\citet{gontijo2020affinity} argue it is the \textit{perceived} distribution shift that needs to be minimised, while maximising the sample vicinity.
Formalising these concepts, they introduce 
augmentation ``diversity'' and ``affinity''.
Diversity is defined as the training loss when learning with artificial samples, while affinity quantifies the difference between the accuracy on original test data and augmented test data, for a reference model.
The latter penalises augmentations that introduce artificial information to which the model is not invariant, \textit{implicitly assuming that training with that information is detrimental to generalisation}. Thus, in contrast to the evaluation methods mentioned above, the artefacts are considered non-negligible when training with distorted data.

In summary, it is currently assumed that the artefacts introduced by changes in the data are negligible when evaluating models, while those introduced when training are important and undesirable. 
These assumptions implicitly shape the community's perception of how machine learning works.
Does the artificial information added by analysis methods not have major side-effects or does it lead to biased results? Conversely, are the artefacts important when training with modified data? Do they cause models to learn better or worse representations? 

We set out to answer these questions. We find that results can be misleading when not accounting for the secondary effects of data manipulation, especially in comparative studies.
Taking an oversimplified example to illustrate this for robustness, we can imagine a binary cat--truck image classification problem and two models: model $A$, which identifies cats solely by the presence of pointy ears and model $B$, which has a more holistic approach. Generally, masking out the ears will cause model $A$ to misclassify cats and we would consider this model not robust to occlusion, while model $B$ will continue to correctly classify them. However, if the ears are covered with a large rectangle that introduces horizontal and vertical edges strongly associated with the truck class, this will cause model $B$ to also misclassify. In this case, because the misclassification is not caused by the absence of a feature but rather by the presence of a distractor, we would still consider model $B$ robust to occlusion although its performance degrades. 
In such a case \cutocc would be unable to distinguish between a model that incorrectly classifies because of lack of information, or because of the presence of confounding artefacts, making it an incorrect proxy for measuring occlusion robustness. Thus, the side-effects of data distortion must be taken into account to create fair evaluation methods.

We start by introducing a measure that indicates the existence of such side-effects which we then use to disprove the negligibility of data distortion artefacts when evaluating models.
We then build on the importance of artefacts to construct empirical counter-examples which disprove common beliefs in the augmentation literature, and highlight the importance of understanding the changes data manipulation introduces. 
Our contributions are as follows:
\begin{itemize}
    \item We introduce a quantity for highlighting the interference of artificially introduced information with a model's learnt representations (Section~\ref{sec:artefacts});
    \item We show that increasingly popular model interpretation and analysis methods are biased (Section~\ref{sec:artefacts});
    \item We propose a fairer alternative for measuring occlusion robustness (Section~\ref{sec:iocc});
    \item We show that in contrast to what is widely assumed, not preserving the data distribution can lead to learning better representations (Section~\ref{sec:MSDA}). 
\end{itemize}

An immediate application of our work is in
robustness evaluation of real-world vision models.
Correctly evaluating the ability of a model to perform in the presence of visual occluders is of paramount importance for safety-critical applications such as self-driving cars and medical diagnosis and intervention.
While the impact of our practical contributions is relevant to the community, we believe combating erroneous research directions is more important for the future development of the field.
Correctly understanding the increasingly popular mixed-sample augmentation is essential for trusting its usage in sensitive applications where the data can be out of distribution. 
Moreover, studying the effects of image distortion can shed further light on what \textit{models} perceive as being within distribution. 
Most importantly, we believe this could set a new direction in capturing the relationship between data and learned representations, which could ultimately play a role in understanding generalisation.

\section{Are Artefacts Negligible when Analysing Classifiers?} \label{sec:artefacts}

In this section we show that artificially introduced artefacts may not be negligible, and distorting data at evaluation time could have side-effects not previously considered.
Specifically, the artefacts can interfere with the representations learnt by the model, which in turn leads to incorrect evaluation.
We highlight this interference by showing that \textit{the distortion can be consistently associated with a particular class in an image classification task}. 
We do so by looking at the increase in misclassifications per predicted category; from the number of incorrect predictions of a model evaluated on modified data, we subtract the incorrect predictions when testing on original data.
If, across multiple runs, there is a significant increase for a specific class, this indicates that the distortion introduces features the model associates with that class.
We refer to this phenomenon as ``data interference''.
By a model ``run'' we refer to a model instance trained with a different seed for the initialisation of weights and for the randomised data augmentation.
For computing the \di index we train 5 different instances of the same model and only consider that data interference occurs when runs consistently display a bias for the same class.

Considering only positive differences, we denote the increase in the percentage of misclassifications for predicted class $c$ for a run $r$ by ${c}^{r}$.
Note that the class is taken to be that predicted by the classifier.
To keep the score within a consistent range across data sets, we scale $c^{r}$ by the number of classes.
We define the Data Interference (\di) index as

\begin{equation*}
    \av[r]{\frac{c_{max}^{r}}{\sum_{c} c^{r}} \, \av[r^{'}]{c_{max}^{r^{'}}} },
\end{equation*}

where $c_{max}$ is that of the class with the highest mean increase across all runs.
The \di index measures the proportion represented by the dominant class weighted by its average increase across runs.
A high index value indicates a sharp increase for a particular class which is consistent across runs. 
We associate this with an overlap between introduced artefacts and learnt representations, thus highlighting the \textit{side effects of distorting data for model evaluation}.

To obtain models with different behaviours in a controlled manner, we make use of Mixed Sample Data Augmentation (MSDA).
Since it is sufficient to identify some common cases in which models are disfavoured, we choose to reduce our environmental impact by restricting the analysis to simple MSDAs that combine images without incurring additional computation time or external models. 
As will be argued in Section~\ref{sec:iocc}, we expect the unfairness to be present in most settings, thus the exact choice of augmentation is irrelevant.
We focus on two popular MSDAs, \mixup and \cutmix~\cite{yun2019cutmix}.
\mixup linearly interpolates between two images to obtain a new training example, while \cutmix masks out a rectangular region of an image with the corresponding region of another image. Besides the aforementioned methods, we also employ \fmix~\cite{harris2020understanding} due to its irregularly shaped masks sampled from Fourier space, which will play an important role in our analysis.
Note that although the masking methods sample the size of the occluding patch from the same distribution, in \cutmix part of the rectangle can be outside the image, which leads to less occluded samples overall.
We refer to models by the augmentations they were trained with and use ``basic'' to label the models trained without MSDA.

Throughout the paper, we do 5 runs of each experiment with PreAct-ResNet18~\cite{he2016identity} as the default architecture.
We include a number of results for BagNet~\cite{brendel2018approximating}, VGG~\cite{Simonyan15}, WideResNet-28-10~\citep{zagoruyko2016wide} and DenseNet-BC-190~\citep{huang2017densely} in the Appendices (e.g. \ref{sup:tiny_jig}, \ref{sup:architectures}) where we also include a cross-architecture analysis.

The main data sets we report results on are \cifar{10/100}~\cite{krizhevsky2009learning}, Tiny~\imagenet\cite{tinyimagenet}, FashionMNIST~\cite{xiao2017fashion}, and \imagenet~\cite{ILSVRC15}.
For \imagenet we use pretrained ResNet-101 models made publicly available by \citet{harris2020understanding}. Note that for \imagenet we do not report results with repeated runs since only one model per augmentation was available.
For full experimental details, see Appendix~\ref{sup:exp_details}. 
We provide the code at \href{https://github.com/AntoniaMarcu/Data-Modification}{https://github.com/AntoniaMarcu/Data-Modification}.

\subsection{Shape Bias} 
\begin{figure}
\vskip 0.2in
    \centering
\includegraphics[width=\linewidth]{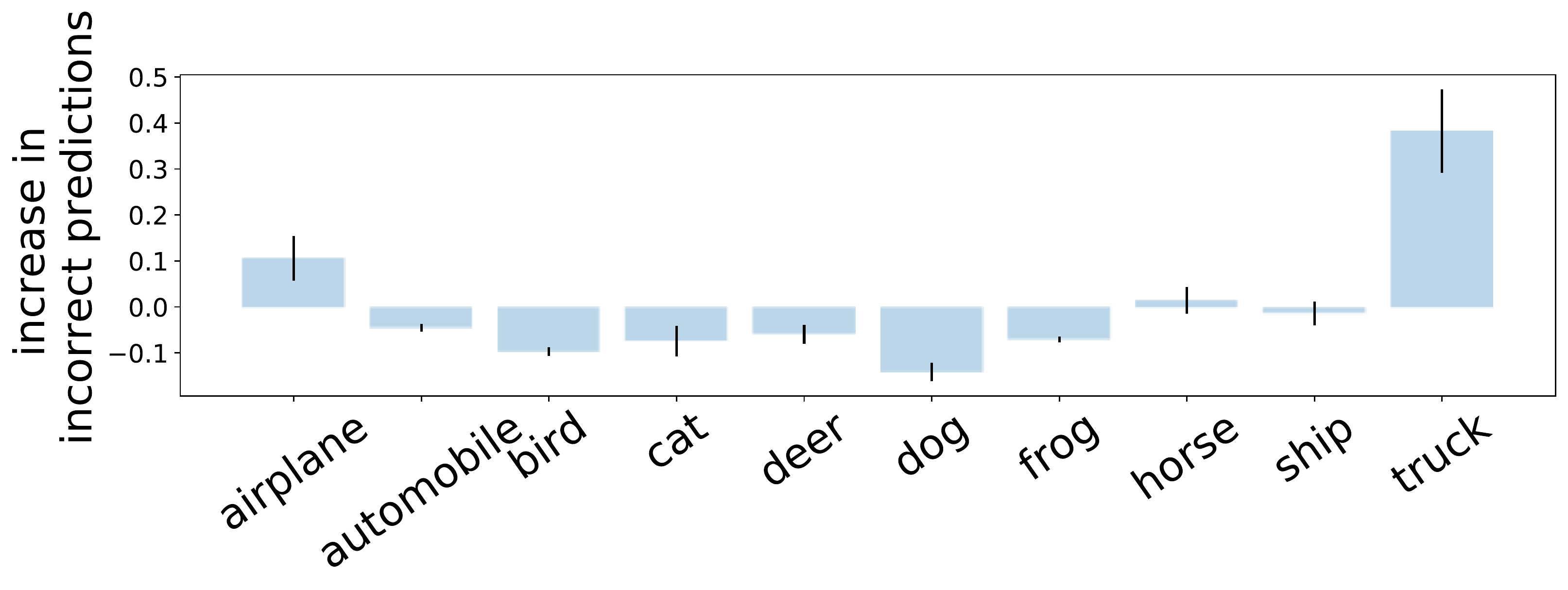}
    \caption{
    Change in incorrect predictions for five basic \cifar{10} models when evaluated on patch-shuffled versus original samples.  
    The models consistently associate patch-shuffled images with the ``truck'' class as indicated by the high increase in incorrect predictions for this class.
    }
    \label{fig:puzzle_truck}
    \vskip -0.2in
\end{figure}

Using the DI measure, we want to show the existence of side-effects that occur when measuring shape bias by the accuracy after patch-shuffling images. This bias makes the patch-shuffling evaluation method unreliable.
For assessing shape bias through sample manipulation, the standard procedure is to choose between dividing the image in 4, 16 or 64 patches to be shuffled.
Since FashionMNIST images are smaller, we choose a $2 \times 2$ grid, 
while for \cifar{10/100}, Tiny~\imagenet  and \imagenet we use a $4 \times 4$ grid.
The results for this experiment are given in ~\cref{tab:sh-tx-bias}.
To provide a sense of scale note that when distorting with Gaussian noise, the basic (trained without MSDA) model has a DI index of $0.09_{\pm 0.02}$ (see Appendix~\ref{sup:randdi}).
There is no fixed threshold that indicates data interference as this would require an objective definition of bias.
Instead, the results are meant to be interpreted in a comparative manner.
As such, we are interested in determining if there exists a clear gap between different models across the same task. We observe that for all four data sets such a gap exists, with the basic and \mixup models generally having high index values. 

Reiterating, a large index indicates that a models tends to associate the features artificially introduced by patch-shuffling with a certain class.
Taking a closer look at the distribution of misclassifications for \cifar{10} we can see that in the case of the basic model this class is ``Truck'' (\cref{fig:puzzle_truck}).
The result is not at all surprising, given that the strong horizontal and vertical edges are highly indicative of this class.
Similar observations can be made for other data sets (Appendix~\ref{sup:wrongPred_shape}).
Thus, we believe the patch-shuffling approach could cause models which are not invariant to strong horizontal and vertical edges to falsely appear to rely more on shape information.
To verify this claim, we first need to check if sensitivity to patch shuffling is necessarily a symptom of shape bias.
Thus, we want to verify if models with similar shape bias can have different DI index values; if so, patch-shuffling would be an unfair basis for evaluation of the shape bias of such models.

\begin{table}[!t]
    \centering
    \caption{DI index for PreAct-ResNet18 on grid-shuffled images for four different types of models. 
    The higher the index, the more a model would be disadvantaged by side-effects caused by the distortion.
    For each data set, we show the results with the lowest average in blue and the ones with the highest average in green.
    Information introduced by shuffling tends to interfere less with 
    \fmix and \cutmix models.
    } \label{tab:sh-tx-bias}
    \vskip 0.15in
    \begin{tabulary}{\linewidth}{lLLLL}
    \toprule
            & basic  & \mixup & \fmix   & \cutmix     \\
     \midrule
    \cifar{10} & $\color{green}{2.82_{\pm 0.44}}$ & $2.40_{\pm 0.59}$ & $0.59_{\pm 0.12}$ & $\color{blue}{0.31_{\pm 0.10}}$\\
    \cifar{100} & $\color{green}{0.99_{\pm 0.27}}$ & $0.88_{\pm 0.24}$ & $0.18_{\pm 0.10}$ & $\color{blue}{0.09_{\pm 0.04}}$ \\
    Fashion & $1.23_{\pm 0.15}$ & $\color{green}{2.42_{\pm 0.92}}$ & $1.06_{\pm 0.23}$ & $\color{blue}{0.68_{\pm 0.11}}$ \\
    Tiny & $\color{green}{1.28_{\pm 1.13}}$ & $0.57_{\pm 0.11}$ & $0.67_{\pm 0.10}$ & $\color{blue}{0.25_{\pm 0.11}}$ \\
\imagenet & $0.82$ & $\color{green}{1.49}$ & $\color{blue}{0.58}$ & $-$\\
    \bottomrule
    \end{tabulary}
    \vskip -0.1in
\end{table}

\textbf{Is a model necessarily more affected by patch-shuffling if it has a higher shape bias?} 
To explore this we use an alternative, well-known, method of measuring shape and texture, which we call the GST approach.
This implies evaluating \imagenet models on the Geirhos Style-Transfer (GST) data set~\cite{geirhos2018imagenet}.
GST contains artificially generated images where the shape belongs to one class and the texture to another.
The data set was specifically designed for shape and texture bias identification and it represents a gold standard in the field. The limitation associated with it is that it can only assess the bias of models trained on data sets with which it is compatible. For this reason, alternative universal methods such as patch-shuffling were proposed.
The data set contains 16 coarse categories that encompass a number of \imagenet classes to which they are mapped.
The bias of the models is given by the accuracy obtained when the label is set to either the shape or texture.
Using this well-known method we want to find models which have similar biases but different DI indices when patch-shuffling.
This would indicate that sensitivity to shuffling is not necessarily linked to increased shape bias, which in turn would mean that models evaluated using patch-shuffling can artificially appear more shape biased.


\begin{table}
\caption{Accuracy of augmentation-trained 
    models on the GST data set when the label is taken to be either the shape or texture. 
    Higher accuracy indicates a stronger bias.
    There is no clear link between masking and low texture bias.} \label{tab:sh-tx-pretrain}
    \vskip 0.15in
    \begin{tabulary}{\linewidth}{lccCC}
    \toprule
    & \multicolumn{2}{c}{\imagenet} &  \multicolumn{2}{c}{Tiny \imagenet} \\
     \midrule
                & Shape & Texture & Shape & Texture \\
    \midrule
    basic    &  $20.31$ & $53.28$ & $10.56_{\pm 0.65}$ & $26.04_{\pm1.77}$\\
    \mixup      & $24.14$ & $60.31$  &  $12.02_{\pm0.33}$ & $27.77_{\pm1.56}$ \\
    \fmix       & $21.25$ & $53.43$ &  $10.40_{\pm 0.39}$ & $19.90_{\pm2.12}$\\
    \cutmix     & ---& ---    & $10.54_{\pm0.38}$ & $23.72_{\pm 2.42}$\\
    \bottomrule
    \end{tabulary}
    \vskip -0.1in
\end{table}

The results in \cref{tab:sh-tx-pretrain} show that the basic model does not have a higher shape bias than masking methods although it has a significantly higher DI index compared to these, as we have seen in~\cref{tab:sh-tx-bias}.
We repeat the same experiment on the Tiny \imagenet data set. 
\citet{geirhos2018imagenet} used WordNet \cite{miller1995wordnet} to map the 1000 categories to the 16 classes of the GST data set. We used the same method to create a mapping between Tiny \imagenet and GST.
A number of \imagenet categories that belong to the 16 higher-level classes of GST are missing. 
For this reason, a poorer overall performance is expected and the results could differ slightly given a better fit between the sets.
We find again no significant correlation between masking augmentation and texture bias. 
\cref{sup_tab:tiny_bag} of Appendix~\ref{sup:tiny_jig} also includes the results for BagNet models, which have smaller receptive fields and so are forced to use more local information.
Even in this case, we find a high DI for the basic model and no difference in texture bias compared to MSDA. 
Thus, a model which is more affected by the side-effects of patch-shuffling is not necessarily more shaped-bias.
In other words, the artefacts introduced by patch-shuffling can cause models to have different accuracies on these distorted images, albeit not differing in their shape and texture bias.

\begin{table*}
    \begin{minipage}[t]{0.6\linewidth}
\centering
    \caption{DI index when occluding with black patches. We show the results with the lowest average in blue and the ones with the highest average in green. For each data set, there exists a non-negligible gap between masking and non-masking training regimes, with models like \mixup and basic being disadvantaged due to higher data interference.} \label{tab:occlusionBias}
    \vskip 0.15in
    \begin{tabulary}{\linewidth}{lllll}
    \toprule
            & basic  & \mixup & \fmix   & \cutmix \\
     \midrule
        \cifar{10}  & $1.25_{\pm 0.17}$ & $0.47_{\pm 0.11}$ & \color{blue}{$0.11_{\pm 0.04}$} & \color{green}{$2.20_{\pm 0.81}$} \\
    \cifar{100}  & \color{green}{$1.24_{\pm 0.35}$} & $0.34_{\pm 0.09}$ & \color{blue}{$0.12_{\pm 0.10}$} & $1.06_{\pm 0.32}$ \\
    FashionMNIST & $0.21_{\pm 0.08}$  & \color{green}{$0.38_{\pm 0.06}$}  & $0.16_{\pm 0.05}$  & \color{blue}{$0.12_{\pm 0.05}$}\\
    Tiny~\imagenet & \color{green}{$0.52_{\pm 0.17}$} & $0.39_{\pm 0.03}$ & \color{blue}{$0.14_{\pm 0.04}$} & $3.46_{\pm 2.45}$ \\
    \imagenet & \color{blue}{$0.50$} & \color{green}{$1.50$} & \color{blue}{$0.50$} & $-$ \\
    \bottomrule
    \end{tabulary}
    
    \end{minipage}
    \hfill
    \begin{minipage}[t]{0.37\linewidth}
    \centering
    \caption{Accuracy on patch-shuffled images. 
    It is considered in the literature that the higher the accuracy, the more texture-oriented the model is.
    Unlike evidenced through the GST method~(\cref{tab:sh-tx-pretrain}), the basic model appears to have a higher shape bias.} 
    \label{pure_acc:shuff}
    \vskip 0.15in
    \strut\vspace*{-\baselineskip}\newline
    \begin{tabulary}{\linewidth}{lCC}
    \toprule
    & \imagenet & Tiny \imagenet \\
    \midrule
    basic & $49.49$ & $13.53_\pm{2.02}$\\
    \mixup & $52.16$ & $17.81_\pm{0.16}$\\
    \fmix & $56.20$ & $34.53_\pm{0.62}$\\
    \cutmix & --- & $42.30_\pm{0.29}$\\
    \bottomrule
    \end{tabulary} 
    
    \end{minipage}
    \vskip -0.1in
\end{table*}

We further confirm this conclusion using the GST approach on the two data sets with which it is compatible.  \cref{pure_acc:shuff} gives the accuracy when patch-shuffling on Tiny~\imagenet and \imagenet. In both cases, for comparable levels of shape bias, different accuracies are obtained.
This confirms that \textit{models can appear to have vastly different shape bias when evaluated on randomly rearranged patches, when in reality their actual shape bias is similar}.
The sensitivity of the patch shuffling approach to artefacts makes it unfair and unreliable measure of shape bias.

\subsection{Occlusion Measurement}

We next want to determine whether the same issue 
applies to occlusion robustness measures. 
We focus on \cutocc, where a rectangular black patch is superimposed on test images and the robustness is given by the resulting accuracy.
There is no standardised distortion when measuring \cutocc, with the size and positioning of the obstructing patch varying between studies. 
Most often in prior art a lack of robustness is noted for large occluders \cite{chun2020empirical, zhong2020random}.
For this reason, we uniformly sample the size of the patch from [0.7,\,1], allowing the occluding patch to lie outside the image (as it is done when augmenting with \cutmix and \cutout~\cite{devries2017improved}).
This allows us to capture both the cases in which either the centre or the border area is masked out but requires a non-uniform distribution to counter for the patches existing outside the image.

\begin{figure*}
\vskip 0.2in
    \begin{minipage}{0.48\linewidth}
    \centering
        \includegraphics[width=\linewidth]{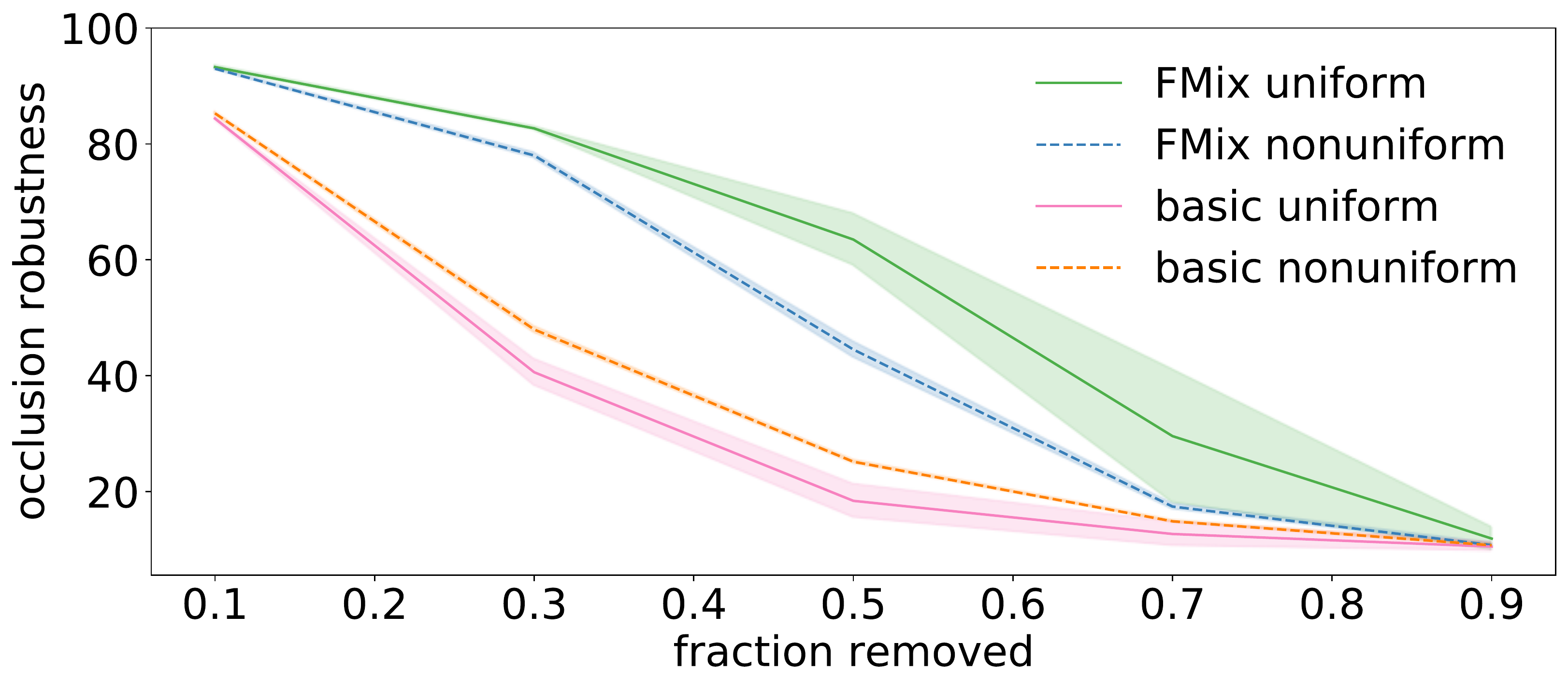}
        \vskip -0.05in
    \caption{\cutocc when occluding with black patches (uniform) and patches taken from other images (nonuniform).
    \cutocc gives different robustness results for the two types of colour patterns, which means that the measure is sensitive to the characteristics of the occluder.}
    \label{fig:mix-nomix-cutocc}
    \end{minipage}
    \hfill 
    \begin{minipage}{0.48\linewidth}
    \centering
    \includegraphics[width=\linewidth]{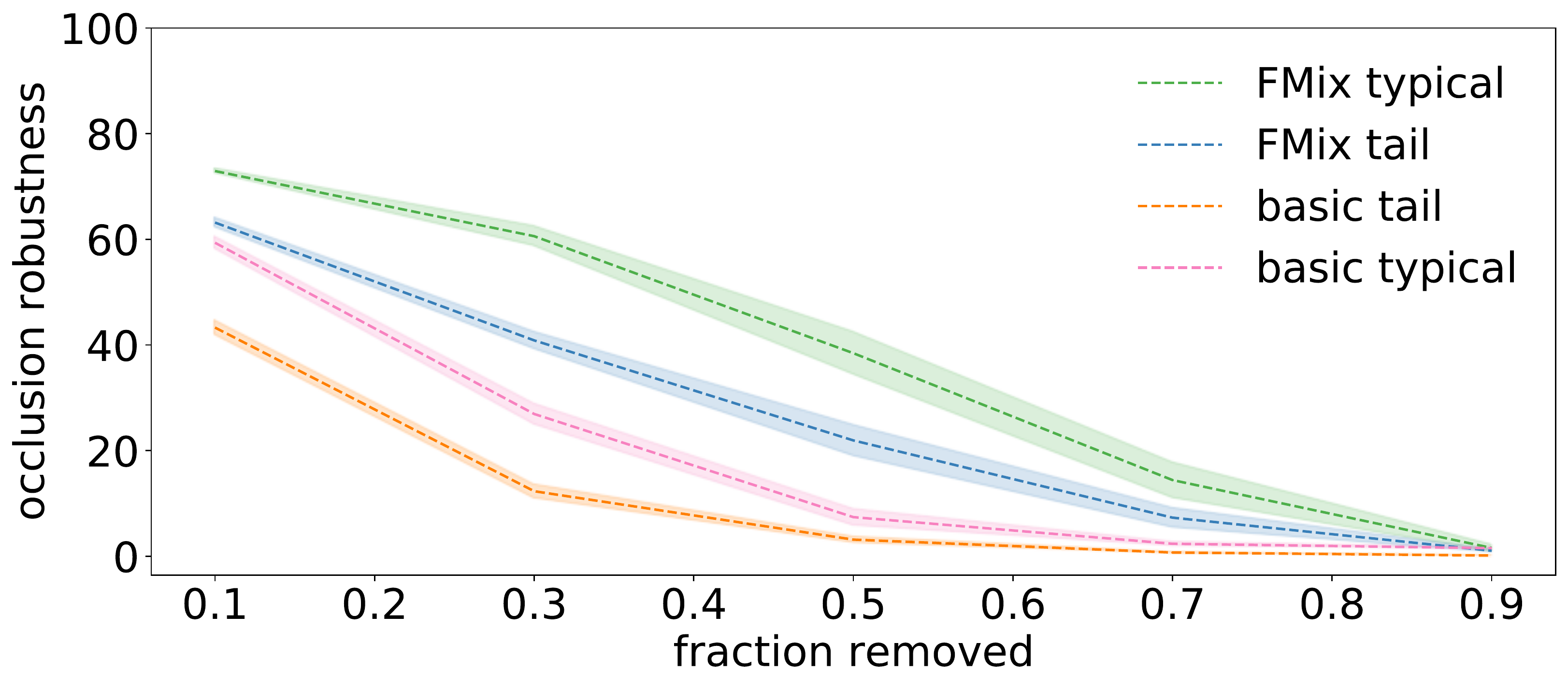}
    \vskip -0.05in
    \caption{ \cutocc for the basic and \fmix models on two subsets of the same data set: tail and typical. \cutocc reports different robustness results on two subsets although the relative drop in accuracy is the same. This is because \cutocc does not abstract away the model's overall performance.}
    \label{fig:tail-typical-cutocc}
    \end{minipage}
    \vskip -0.1in
\end{figure*}

We perform the same experiment as before, where the \di index is now measured when testing on rectangle-occluded images and present results in 
\cref{tab:occlusionBias}.
Once again, we can observe a significant gap in the DI index for each data set. 
This indicates that some models will again be disadvantaged. 
We find that data interference is present for different architectures, 
when overlapping patches from external images or using differently shaped masks (Appendix~\ref{sup:more_cutocc_DI}).
Thus, the result of \cutocc and its variants is highly dependent on the problem at hand.
Just as for randomly shuffling tiles, by occluding images using a particularly shaped patch, one implicitly measures a model's affinity to certain features.
In other words, models for which a distinctive feature is artificially introduced when performing \cutocc are going to be penalised by this method even though the feature is perfectly valid in a normal classification setting. 
This deems such methods inappropriate for fairly assessing robustness and texture bias.

To demonstrate that \cutocc is sensitive to the occluder specifics, we superimpose patches taken from a different data set and compare the results to those obtained when occluding with black patches only. For clarity, \cref{fig:mix-nomix-cutocc} presents the results for the least and most robust models.
The experiment indicates that it is sufficient to simply change the colour pattern of the patch for \cutocc to give different results outside the margin of error.

Moreover, this is not the only limitation of \cutocc. Another problem that occurs when looking purely at post-masking accuracy is that weaker models would erroneously appear less robust.
We show this by reversing the problem: we evaluate the same model on two different subsets of the \cifar{100} data set: typical and tail images as categorised by \citet{feldman2020neural}. 
They consider a train-test sample pair to belong to the tail of the data distribution if the test sample is correctly classified when a model is trained with the train sample, and incorrectly without it.
\cutocc would indicate that models are significantly more robust to occluding typical examples.
However, a closer analysis makes us doubt this conclusion.
The raw accuracy on both train and test data for tail examples is lower than for the typical ones.
In fact, the performance when occluding images decreases at the same \textit{rate} for the two subsets, indicating similar robustness. 

We have seen that the results obtained using \cutocc are sensitive to a number of factors.
To better reflect how much information can be hidden from a model without affecting its performance, a fair alternative should be invariant to the model's goodness-of-fit and to the characteristics of the occluder.
Concerning the latter, a related observation was made by \citet{hooker2019benchmark} who note the pitfalls of manipulating data to determine feature importance.
They point out that when simply superimposing uniform patches over image features, it is difficult to assess how much of the reduction in accuracy is caused by the absence of those features and how much is due to images becoming out of distribution.
To address this, the most important features identified by an estimator are masked out both on train and test data, closing the gap between the two sets. 
\citeauthor{hooker2019benchmark} then train and evaluate models on the newly generated images.
Unlike with interpretability, the subject of occlusion robustness studies is the model itself, which makes training with a modified version of the data inviable.
In the following section we explore ways of overcoming this bias when measuring occlusion robustness.

\section{What are Fairer Alternatives?} \label{sec:iocc}

\begin{figure}
\vskip 0.2in
    \centering
    \includegraphics[width=0.9\linewidth]{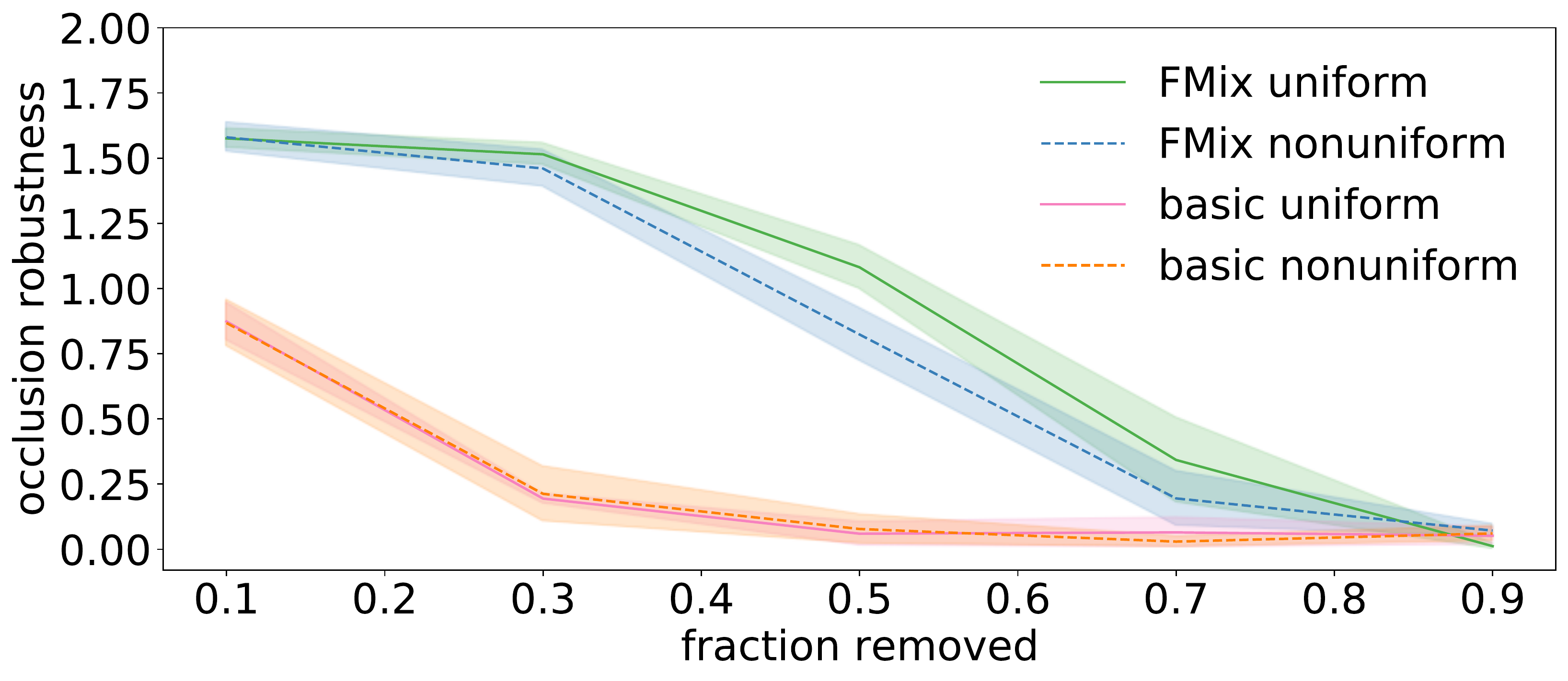}
    \caption{ \iocc when occluding with black patches (uniform) and patches taken from other images (nonuniform).
    Unlike \cutocc, our measure gives consistent results when varying the colour pattern of the occluder.
    }
    \label{fig:mix-nomix-iocc}
    \vskip -0.2in
\end{figure}

We propose a simple, carefully defined measure that aims to \textit{decouple the model's edge bias and generalisation ability from the occlusion robustness.}
We refer to our measure as ``interplay occlusion'' (\iocc).
Interplay occlusion reflects the change in the interplay between performance on seen and unseen data. Formally,
\begin{equation}
    \iocc_{p} = \left| \frac{ \mathcal{A}{(\mathcal{D}^{p}_{train})}- \mathcal{A}{(\mathcal{D}^{p}_{test})}}{\mathcal{A}{(\mathcal{D}_{train})}- \mathcal{A}{(\mathcal{D}_{test})}} \right|,    
    \label{eq:iocc}
\end{equation}
where $\mathcal{A}(\mathcal{D})$ denotes the accuracy on a given data set $\mathcal{D}$, and $\mathcal{D}^{p}$ is the data set resulting from removing $p\%$ pixels of each image.
The intuition is that on training data, robust models are less sensitive to the artefacts of the occlusion policy for small levels of occlusion, resulting in a large difference in accuracy from that on unseen data.
The performance of both train and test gets close to random as the proportion of occluded data approaches 90\%. We expect the gap to fall off quicker for less robust models. 
This change in interplay is taken with respect to the generalisation gap of the model, such that the quality of the model fit in itself does not interfere with the robustness measure. 

A number of factors have to be considered when choosing a masking method for computing $\mathcal{D}^{p}$, such as the number of contiguous components or the amount of salient information masked out.
In this paper we choose to generate masks using Grad-CAM \cite{selvaraju2017grad}. 
Since methods could be sensitive to either occluding contextual or core information, we get an average performance by choosing with equal probability between occluding the \textit{most} or \textit{least} salient $p\%$ pixels for each batch.
It must be noted that this method implicitly assumes there could be multiple occluders, which has the downside of incurring a higher computational cost.
We also experiment with using rectangular or Fourier-sampled masks and conclude that although random masking makes the process noisier, the exact choice of masking method is of secondary importance as long as the occluder's granularity is accounted for.
Appendix~\ref{sup:approximating} provides discussion and results on these alternative instances of \iocc.
Throughout this section, for a fair comparison, when measuring \cutocc we do not allow the obstructing patch to lie outside the image. As such, the fraction removed is exact.

\begin{figure}[t]
\vskip 0.2in
    \centering
    \includegraphics[width=0.9\linewidth]{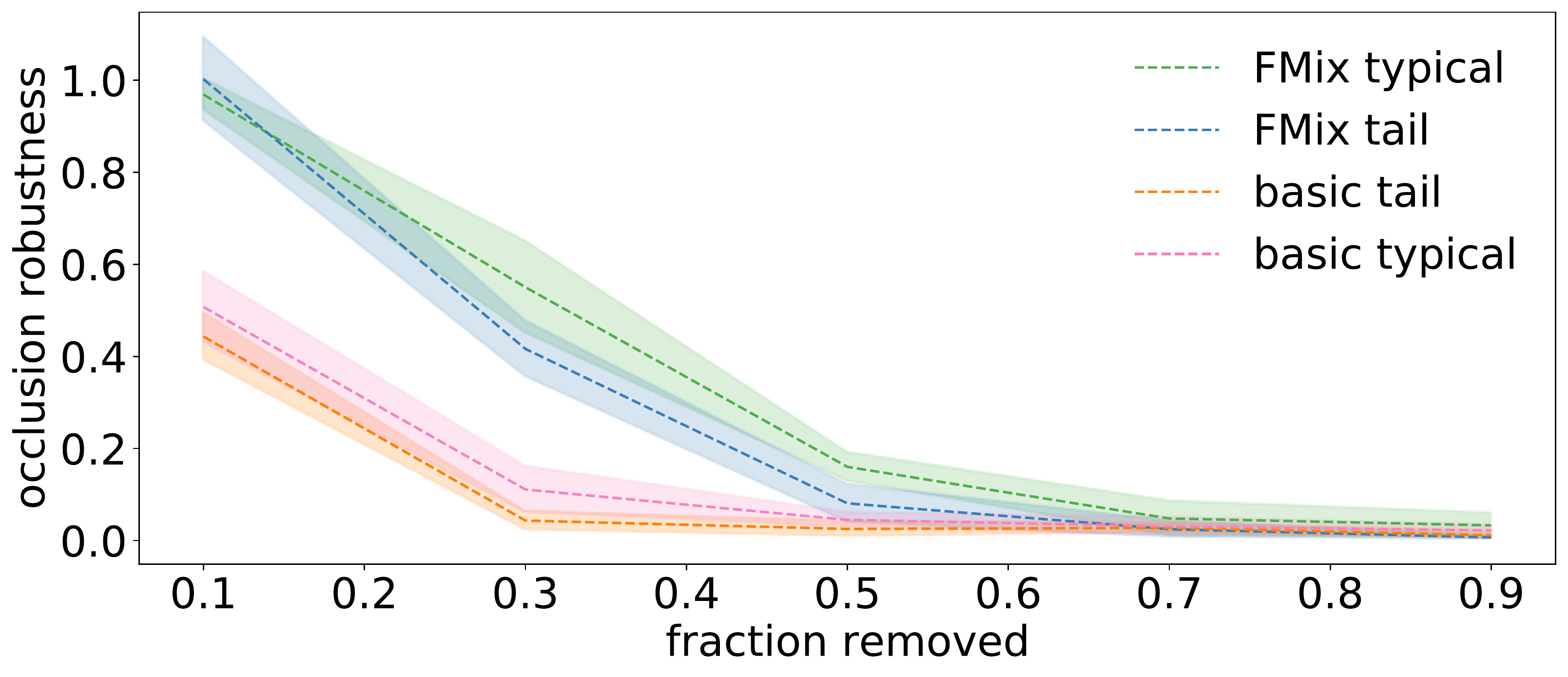}
\caption{\iocc for basic and \fmix models on tail and typical data. As opposed to \cutocc, measuring robustness with \iocc reflects the overlapping robustness level of the models on the two subsets.}
    \label{fig:tail-typical-iocc}
    \vskip -0.2in
\end{figure}

\begin{table}[t]
    \centering
    \caption{DI index and occlusion robustness for models trained on \cifar{10} when obstructing 30\% of the image pixels with non-uniform patches. Given in bold is the closest result to that of RM.
    The higher DI index of RM compared to the other masking methods is causing \cutocc to evaluate the robustness of RM as closer to \mixup, rather than \fmix or \cutmix.
    }\label{tab:3msks}
    \vskip 0.15in
    \begin{tabular}{ll|ll}
    \toprule
         & DI index & \cutocc & \iocc \\
        \midrule
        basic & $1.67_{\pm0.17}$ & $47.97_{\pm0.52}$ & $0.21_{\pm0.10}$ \\
        \mixup & $0.98_{\pm0.21}$ & $\bm{58.65_{\pm1.01}}$ & $0.57_{\pm0.18}$\\
        \cutmix & $0.14_{\pm0.08}$ & $76.56_{\pm 6.36}$ & $\bm{1.09_{\pm 0.17}}$\\
        \fmix & $0.15_{\pm0.01}$ &$78.00_{\pm 0.45}$ & $1.46_{\pm 0.07}$\\
        RM & $0.39_{\pm0.05}$ & $60.79_{\pm 5.03}$ & $1.20_{\pm 0.23}$\\
    \bottomrule
    \end{tabular}
    \vskip -0.1in
\end{table}

Assessing the correctness of such a measure is difficult in the absence of a baseline. 
We construct varied experiments to attest the validity of our method, highlighting the two important limitations of \cutocc that our approach addresses: sensitivity to the specifics of the occluding patch and to model's generalisation performance.
For the curiosity of the reader, we include various additional experiments and analyses in Appendix~\ref{sup:further_results_iocc}, and focus here on the key results.
We start by performing the same experiment as for \cutocc measurement where we occlude using black patches and patches taken from a different data set.
As shown in \cref{fig:mix-nomix-iocc}, \iocc using uniform occluders gives similar results to its non-uniform version and thus provides a consistent model evaluation.
We then verify that \iocc is invariant to the model's overall performance by evaluating on the same two subsets of the \cifar{100} data set on which we evaluated \cutocc. The results obtained using \iocc on tail and typical data are within each other's margin of error (\cref{fig:tail-typical-iocc}). Thus, \iocc abstracts away from the colour of the occluder as well as the model's generalisation performance.

In addition to not being sensitive to the colour pattern of the patch, a fair measure must also be invariant to the shape of the patch.
To empirically confirm \iocc reduces the importance of edge information, we aim to obtain a model that is robust to occlusion, but at the same time has a high DI index (it is sensitive to edge information).
To this end, we create a variation of \fmix, Random Masks (RM), where at the beginning of the training process three masks are randomly sampled from Fourier space.
For each batch, one of the three is chosen uniformly at random. 
When masking with patterned patches the RM model has a higher DI index than \fmix, as desired.
\cref{tab:3msks} gives results for a fraction of 0.3 pixels covered by a non-uniform occluder.
Our measure reflects the robustness of training with RM, situating it close to other masking methods. 
On the other hand, because \cutocc implicitly penalises models with high DI index, according to this measure RM appears almost as sensitive to occlusion as \mixup. 
\cref{fig:3msks} shows results for a wider range of fractions.

The unbiased nature of \iocc could lead to better understanding and development of training procedures.
Equally important, \iocc has applicability to real-world vision system deployments where no prior knowledge exists about the possible shapes of the obstructions. 
While this aspect of generality is the strength of our approach, it must be stressed that when there exists a limited set of known possible occluders, evaluating robustness specifically to them could be safer.
For example in an industrial setting or a clinical environment there could be a certain set of objects that could interfere with the subject of interest.
We do not propose a universal solution, but rather suggest an alternative to the biased approach for the common scenario in which the environment is not controlled and little is known about all the potential occluders. 

The strength of the bias will depend on the data in question and some applications will be more affected than others. 
We have seen that for natural images this bias does exist.
To confirm that we have not just identified isolated cases, we remove the class that has the highest increase in mispredictions and retrain the models on the remaining classes. 
We find that the bias is again present (\cref{tab:di_exclude}), but with respect to another class. For example, in the case of \cifar{10}, after removing the ``Truck'' class, models mispredict rectangle-occluded images as ``Boat'' (Appendix~\ref{sup:removeDominnat}).
Thus, the edge artefacts are likely to interfere with learnt representations since they are such fundamental features.
From an evaluation perspective, as we have seen, this impacts assessment methods and must be accounted for.
From the training perspective, such a widespread data interference would indicate a large perceptual shift when performing MSDA.
However, large perceptual shifts are believed to have a negative effect on generalisation, while MSDAs are known to improve performance. Is then the perceptual shift caused by artificially introduced information really detrimental to learning?
In the following section we investigate the impact of artefacts on training.


\begin{table}[!tb]
    \centering
    \caption{Augmentation comparison on \cifar{10}. We consider two variants when calculating diversity. Diversity computes the cross-entropy loss using the label of the majority class, whilst MixDiversity is a linear combination of the two cross-entropy losses.
    In the literature it is argued that higher affinity and higher diversity are desirable, but we find that this is not necessarily the case.}
    \label{tab:affinity}
    \vskip 0.15in
    \begin{tabulary}{\linewidth}{llll}
    \toprule
    & Affinity & Diversity & MixDiversity\\
    \midrule
    \mixup  & $\mathbf{-12.58_{\pm 0.14}}$ & $\mathbf{0.41_{\pm 0.01}}$ & $\mathbf{0.84_{\pm 0.00}}$\\
    \fmix & $-25.55_{\pm 0.26}$ & $0.34_{\pm 0.01}$ & $0.65_{\pm 0.00}$\\
    
    \bottomrule
    \end{tabulary}
    \vskip -0.1in
\end{table}

\begin{table}[t!]
    \centering
    \caption{DI index measured for non-uniform occlusion when training without the class with highest increase in incorrect predictions.
    A gap can be noted in this case as well, \mixup being more affected by the distortion's artefacts as evidenced by the higher \di index.
    This result supports the idea that data interference is not specific to peculiar cases, affecting model evaluation more broadly.
    }
    \label{tab:di_exclude}
    \vskip 0.15in
    \begin{tabulary}{\linewidth}{lll}
    \toprule
    & \cifar{10} & \cifar{100}\\
    \midrule
    \mixup  & $0.39_{\pm 0.15}$ & $1.22_{\pm0.19}$\\
    \fmix & $0.08_{\pm 0.06}$ & $0.50_{\pm 0.21}$\\
    \bottomrule
    \end{tabulary} 
    \vskip -0.1in
\end{table}

\section{Is the Magnitude of the Distribution Shift Important?} \label{sec:MSDA}

Note once again that in this paper we are interested in data augmentation in the classical sense, where distortions are simply used to enlarge the data set. Augmentation has also been used in contrastive representation learning~\citep[e.g.][]{bai2021directional,chen2020simple,he2020momentum}, but we are concerned with scenario in which the model learns to perceive distorted and original data as belonging to the same distribution.
In this case, it was traditionally believed that a good augmentation should have minimal distribution shift. 
However, increasingly many approaches propose heavy distortions~\citep[e.g.][]{yun2019cutmix, summers2019improved}, questioning the traditional view.

Most recently, it has been argued that it is the \textit{degree of shift as perceived by the model} that determines augmentation quality.
\citet{gontijo2020affinity} propose to measure this shift by the difference between the performance of the model when presented with original test data and augmented test data.
They refer to this gap as ``affinity''.
We show that \textit{the magnitude of the distribution shift does not determine augmentation quality} of MSDA.
We start with the perceptual gap of training with MSDA, as proposed in \citet{gontijo2020affinity}. 
Subsequently, we address the gap in the wider sense, as is often sought in prior art.
We first argue that high affinity and high diversity are not necessarily desirable.
Indeed, we find \fmix, a better performing train-time augmentation in terms of impact on generalisation performance, to have both lower affinity and lower diversity than \mixup (\cref{tab:affinity}). 
For diversity, we compute the cross-entropy loss where the label is taken to be that of the majority class.
Similar results are obtained with the \mixup loss, where a weighted average of the true labels is taken. 

While intuitively for a high level of affinity, high diversity could correspond to better methods, the converse does not hold.
We argue this is because affinity is rather an analysis of the learnt representations of the reference model and cannot give an insight into the quality of the augmentation or its effect on learning. 
The limitations of affinity are intimately linked to those of \cutocc.
We have seen in Section~\ref{sec:iocc} that the bias of the basic model is present not only when obstructing an image with a uniform patch, but also when mask-mixing.
As such, an augmentation will have a lower affinity if it introduces artefacts that could otherwise lead to learning better representations when used in the training process.
We believe this issue extends to other approaches that aim to motivate the success of MSDA through reduced distribution shift.
Henceforth, we focus on bringing further supporting evidence that \textit{the importance lies in the invariance introduced by the shift and its interaction with the given problem rather than its magnitude}.

\subsection{If it is not the Magnitude that Matters, is it the Direction?}
\label{sec:bias_matters}

\begin{table*}[!t]
    \caption{Accuracy on \cifar{10} (left) and \cifar{100} (right) upon mixing with samples from a different data set. The baseline is the accuracy when training with a single data set using the reformulated objective. \cifar{110} refers to mixing with \cifar{100} when training on the \cifar{10} problem and vice-versa. 
    Mixing \cifar{10} images with \cifar{100} is beneficial for the former (increases the average performance by 0.51\%) and detrimental for the latter (decreases the average performance by 2.32\%).}
    \label{tab:other}
    \vskip 0.15in
    \centering
    \begin{tabulary}{\linewidth}{lLLLLLLL}
    \toprule
    & \mixup & \fmix & \cutmix & & \mixup & \fmix & \cutmix\\
    \midrule
    baseline  & $94.18_{\pm0.34}$ & $94.36_{\pm 0.28}$  & $94.67_{\pm 0.20}$ & &$74.68_{\pm 0.37}$  & $75.75_{\pm 0.31}$ &  $74.19_{\pm 0.50}$\\ 
    \cifar{110} & $94.70_{\pm 0.27}$ & $94.80_{\pm 0.32}$& $94.66_{\pm 0.12}$ & & $72.36_{\pm1.04}$ & $74.80_{\pm 0.55}$ & $74.47_{\pm 0.39}$\\ 
    Fashion & $92.28_{\pm0.28}$ & $95.03_{\pm 0.10}$ & $94.61_{\pm 0.19}$ & & $66.40_{\pm 1.86}$ & $74.46_{\pm0.57}$ & $74.06_{\pm 0.28}$\\
    \bottomrule
    \end{tabulary}
    \vskip -0.1in
\end{table*}

We propose the study of introduced bias as a more informative research direction.
We use the term ``bias'' to refer to a drift in the learnt representations introduced by the change in the training procedure.
In this section we perform inter-dataset augmentation and show that 
by distorting the data distribution by 
a similar \textit{magnitude}, we obtain two opposing results.
This suggests that it is the \textit{direction} of the introduced bias that is important for understanding the impact of augmentation.

We use the reformulated objective setting \cite{huszar2017, harris2020understanding}, where targets are not mixed and the mixing coefficient is drawn from an imbalanced Beta distribution. 
This allows us to apply MSDA between data sets. 
Thus, for training a model on a data set, we use an additional one whose targets will be ignored.
As an example, a model that is learning to predict \cifar{10} images will be trained on a combination of \cifar{10} and \cifar{100} images, with the target of the former.
This scenario breaks the added correlation between training examples. 
\cref{tab:other} contains the results of this experiment, showing that an accuracy similar to or better than that of regular MSDA can be obtained by performing inter-dataset MSDA.
This invalidates previous arguments that the power of \mixup comes from causing the model to act linearly between samples~\cite{zhang2018mixup}, and calls for a broader explanation of its success beyond that of pushing the examples towards their mean~\cite{carratino2020mixup}.
Another observation is that for \fmix and \mixup, introducing elements from \cifar{100} when training models on the \cifar{10} problem does not harm the learning process.
The reciprocal, however, does not hold.
Hence, the ``distribution shift'' is more intimately linked to the problem at hand and aiming to characterise an augmentation based on the distance from the original distribution is a limiting approach, especially when the distance is measured as perceived by a reference model.

We believe an explanation is that the artefacts created when putting together images from \cifar{10} with those of \cifar{100} could introduce information that makes the separation of the 10 classes easier. 
However, if the information happens to interfere with a feature that is important for separating the \cifar{100} categories, the performance could degrade on this data set. 
This single experiment is not sufficient to draw any general conclusions. 
However, it does show that shifting two distributions by a similar amount can have different effects on the model performance.
Thus, \textit{the direction of the introduced bias could be more important than its magnitude}.
While some level of data similarity has to be preserved when performing MSDA, it is far from being the objective of such data-distorting approaches.

We have highlighted that the shift in learnt representations can lead to better models and simply quantifying the distribution shift can be misleading.
An open question remains: How can we better capture the bias that is introduced and measure its quality? We believe understanding how a relatively small change in the data distribution impacts learnt representations could lead the way to characterising the relationship between data and model generalisation. 

\section{Conclusions}

Studying the side-effects of data distortion is crucial for building fair evaluation methods, better training procedures and even understanding image data as a whole. 
In this paper we show that the artificial information introduced by visual distortion must be accounted for when evaluating models, and it should be studied and exploited during training.
A limitation of previous studies that aim to explain the success of MSDA is the focus on trying to argue similarity with original data, rather than explaining the bias that is introduced. 
Correctly interpreting the bias is important not only for making the models trustable, but also for injecting more informed prior knowledge in future applications.
Beyond their practical benefits, we believe MSDAs have the potential to help characterise the interplay between data and learnt representations.
We hope this paper encourages better practice when dealing with all forms of data distortions.


\section*{Acknowledgements}

A.M. is funded by the EPSRC Doctoral Training Partnership (EP/N509747/1). The authors acknowledge the use of the IRIDIS High-Performance Computing Facility, the ECS Alpha Cluster, and associated support services at the University of Southampton in the completion of this work.
The authors would like to thank Jonathon Hare for the support and helpful discussions.

\bibliography{References}
\bibliographystyle{icml2022}

\newpage
\appendix
\onecolumn
\section{Experimental Details} \label{sup:exp_details}
Throughout the paper, we use PreAct-ResNet18~\citep{he2016identity} models, trained for 200 epochs with a batch size of 128.
For the MSDA parameters we use the same values as~\citet{harris2020understanding}.
All models are augmented with random crop and horizontal flip and are averaged across 5 runs.
We optimise using SGD with 0.9 momentum, learning rate of 0.1 up until epoch 100 and 0.001 for the rest of the training. 
This is due to an incompatibility with newer versions of the PyTorch library of the official implementation of~\citet{harris2020understanding}, which we use as a starting point for model training. 
However, the difference in learning rate schedule between our work and prior art does not affect our findings since we are not introducing a new method to be applied at training time.
In our case, it is sufficient to show that the bias exists in at least one configuration. 
For the analysis we also used adapted code from~\citep{carlucci2019domain} for patch-shuffling.
The models were trained on either one of the following: Titan X Pascal, GeForce GTX 1080 Ti or Tesla V100. 
For the analyses, a GeForce GTX 1050 was also used. The average training time was less than two hours, with the exception of model trained on Tiny-\imagenet, which took around 10 hours to run.

\subsection*{Training Models}
The code for model training is largely based on the open-source official implementation of \fmix, which also includes those of \mixup, \cutout, and \cutmix.
For the experiment where we use the reformulated objective to combine data sets, instead of mixing with a permutation of the batch, as it is done in the original implementation of the mixed-augmentations, we now draw a batch form the desired data set. 
To ensure a fair comparison, for the basic we also perform inter-batch mixing.

\subsection*{Evaluating Robustness}

For the \cutocc measurement, we modify open-source code to restrict the occluding patch to lie withing the the margins of the image to be occluded. This is to ensure that the mixing factor $\lambda$ matches the true proportion of the occlusion.
For \iocc, the implementation of Grad-CAM is again adapted from publicly available code.
With both methods, we evaluate 5 instances of the same model and average over the results obtained.

The added computation time of \iocc over the regular \cutocc for a fixed occlusion fraction is that of performing Grad-CAM on train and test data, as well as evaluating on the latter.
With a batch size of 128, this takes under half an hour. 

\section{Analysis of Wrong Predictions} 

\subsection{Alternative Index} \label{sup:alt_idx}
Table~\ref{tab:sup-sh-tx-bias} gives the worst-case DI index where we replace $c_{max}$ in Equation 1 by the maximum increase across the runs.
As per the original formulation, we note that the masking methods lead to models which are less sensitive to the artefacts resulted after patch-shuffling.


\begin{table}[h]
    \centering
    \caption{Alternative DI index for PreAct-ResNet18 on grid-shuffled images for four different types of models. Again, a bias can be noted for all considered data sets.} \label{tab:sup-sh-tx-bias}
    \vskip 0.15in
    \begin{tabulary}{\linewidth}{lLLLL}
    \toprule
            & basic  & \mixup & \fmix   & \cutmix     \\
     \midrule
    \cifar{10} & $3.52_{\pm 0.56}$ &  $3.31_{\pm 0.82}$ & $0.76_{\pm 0.16}$  & $0.43_{\pm 0.13} $ \\
    \cifar{100} & $1.40_{\pm 0.38}$ &$1.09_{\pm 0.29}$ &
$0.38_{\pm 0.21}$ &$0.16_{\pm 0.08}$ \\
    Fashion & $1.56_{\pm 0.39}$ & $3.57_{\pm 1.35}$ & 
$1.65_{\pm 0.35}$ & $0.82_{\pm 0.13}$ \\
    Tiny & $3.01_{\pm 0.48}$ & $2.24_{\pm 0.30}$ & $2.34_{\pm 1.86}$ & $11.45_{\pm 10.54}$  \\
    \imagenet & $0.82$ & $1.49$ & $0.58$ & $-$\\
     \bottomrule
    \end{tabulary}
    \vskip -0.1in
\end{table}

\subsection{DI Confidence Evaluation and Gaussian Noise}
\label{sup:randdi}

Given the high standard deviation of our index, as a sanity check want to determine what the chances are that the \di we measured is a result of chance.
For this, we compute the \di index where each misclassified example gets randomly assigned to one of the classes, repeating this experiment $10^{5}$ times.
We obtain an average DI of 0.10 with a standard deviation of $4 \times 10^{-5}$.
Thus, the result we obtain is not by chance and the index captures a real phenomenon.

To make our baseline more challenging, we then randomly assign \textbf{all} misclassified examples to one of the classes and compute the index across 5 iterations.
This is so as to simulate the case where each specific model instance consistently associates artefacts with a certain class, but the class is not necessarily consistent across runs.
We perform this experiment $10^{5}$ times and obtain a \di of $0.10_{\pm 0.30}$.

Lastly, we compute the \di index for distorting with Gaussian noise.
We distort the images after the image normalisation step, at which point the pixel values are between 0 and 1.
For each image, we uniformly sample the standard deviation of the Gaussian noise from the interval $[0, 0.1]$, while the mean we set to 0.
For the basic model, we obtain a \di value of $0.09_{\pm 0.02}$.
This is significantly lower than the index we observe in the case of patch-shuffling or rectangular occlusion.

\subsection{Varying the Grid Size} \label{sup:dif_grid_size}

Table~\ref{tab:alt_grid_size} gives the results obtained when varying the number of image tiles to be randomly rearranged. We observe that data interference appears for different grid sizes.

\begin{table*}[ht]
    \centering
    \caption{DI index for alternative grid sizes. Gaps can be identified for various grid sizes, with more pronounced differences for finer-grained grids.}
    \label{tab:alt_grid_size}
    \vskip 0.15in
    \begin{tabulary}{\linewidth}{lLLLLL}
    \toprule
            & & basic  & \mixup & \fmix   & \cutmix  \\
    \midrule
    \multirow{2}{*}{\cifar{10}} & $2 \times 2$ & $0.61_{\pm 0.24}$ & $0.56_{\pm 0.33}$ & $0.19_{\pm 0.14}$ & $0.12_{\pm 0.06}$ \\
                & $8 \times 8$ & $6.41_{\pm 0.55}$ & $6.95_{\pm 1.96}$ & $2.75_{\pm 1.46}$ & $1.41_{\pm 1.15}$ \\
    \midrule
    \multirow{2}{*}{\cifar{100}} & $2 \times 2$ & $1.03_{\pm 0.29}$ & $0.46_{\pm 0.14}$ & $0.21_{\pm 0.14}$ & $0.12_{\pm 0.07}$ \\
                & $8 \times 8$ & $9.16_{\pm 6.15}$ & $3.10_{\pm 4.59}$ & $1.62_{\pm 0.89}$ & $0.65_{\pm 0.50}$ \\
     \midrule

    \multirow{2}{*}{Tiny~\imagenet} & $8 \times 8$ & $5.76_{\pm 6.61}$ & $5.73_{\pm 3.82}$ & $2.49_{\pm 1.38}$ & $0.60_{\pm 0.69}$ \\
    & $16 \times 16$ &  $44.01_{\pm 36.47}$& $14.06_{\pm 14.63}$ &
$11.94_{\pm 17.79}$& $1.86_{\pm 1.98}$ \\
    \midrule
    \multirow{2}{*}{\imagenet} & $16 \times 16$ & $0.72$ & $1.38$ & $0.55$ & $-$ \\
    & $64 \times 64$ & $4.89$ & $41.16$ & $12.77$& $-$\\
    \bottomrule
    \end{tabulary}
    \vskip -0.1in
\end{table*}

\subsection{Patch-shuffling} \label{sup:wrongPred_shape}

\begin{figure*}[h]
\vskip 0.2in
    \centering
    \includegraphics[width=0.8\linewidth]{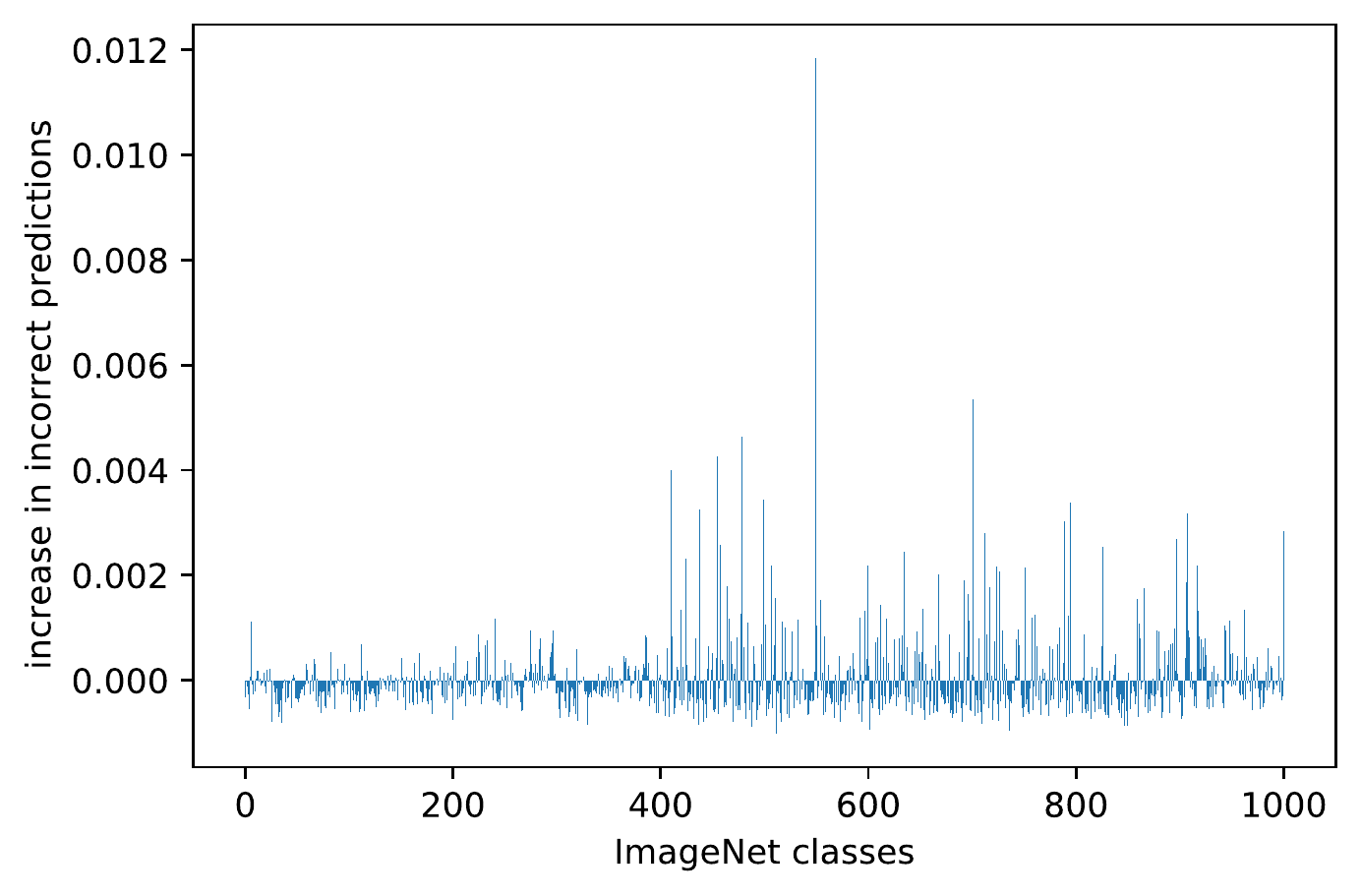}
    \caption{Difference between the number of times a class was wrongly predicted when presented with regular \imagenet samples and patch-shuffled data.} 
    \label{fig:imnet}
\end{figure*}

We look at the classes which have the highest increase in incorrect predictions and note that their shapes are characterised by strong horizontal and vertical edges.
For example, on \cifar{100}, varying the grid size between $2 \times 2$, $4 \times 4$ and $8 \times 8$ gives ``Lamp'', ``Bus'' and ``Table'' as dominant $c_{max}$ classes, while the model trained on Fashion MNIST with the standard procedure tends to predict grid-shuffled images as ``Bag''.
Figure~\ref{fig:imnet} shows that on \imagenet, the basic model tends to wrongly identify the patch-shuffled images as belonging to class ``Envelope''.

\subsection{Increase in Wrong Predictions for the Shape-biased Models}

We attempted to train a model on the GST data set using the shape labels. 
Although we have identified a bias, upon a closer inspection we observed that all model instances were heavily overfitting. 
This is because the GST data set is meant as an evaluation data set, only containing 10 unique shape examples for each class.
Due to time constraints, we could not further explore settings that would give sufficiently well-performing models on which the bias evaluation could be trusted.
Nonetheless, this remains an interesting future direction to explore.

\textbf{Can we propose a fairer shape bias evaluation starting from patch-shuffling?}

The shape-texture issue is far more complicated
and subtle than occlusion. We would argue that the definition of shape and texture on which the patch-shuffling method is based is flawed as it locally maintains both the shape and the texture information. Although it is mostly the texture that is preserved in the patches, edge information is often preserved as well (e.g. a cat’s pointy ears might be fully preserved in a patch). Thus, great care must be taken to make sure that a correct and reliable evaluation of the bias can be constructed with patch-shuffling.

\subsection{\cutocc} \label{sup:more_cutocc_DI}
\begin{figure*}[h]
    \begin{minipage}[t]{0.49\linewidth}
        \centering
        \includegraphics[width=0.75\linewidth]{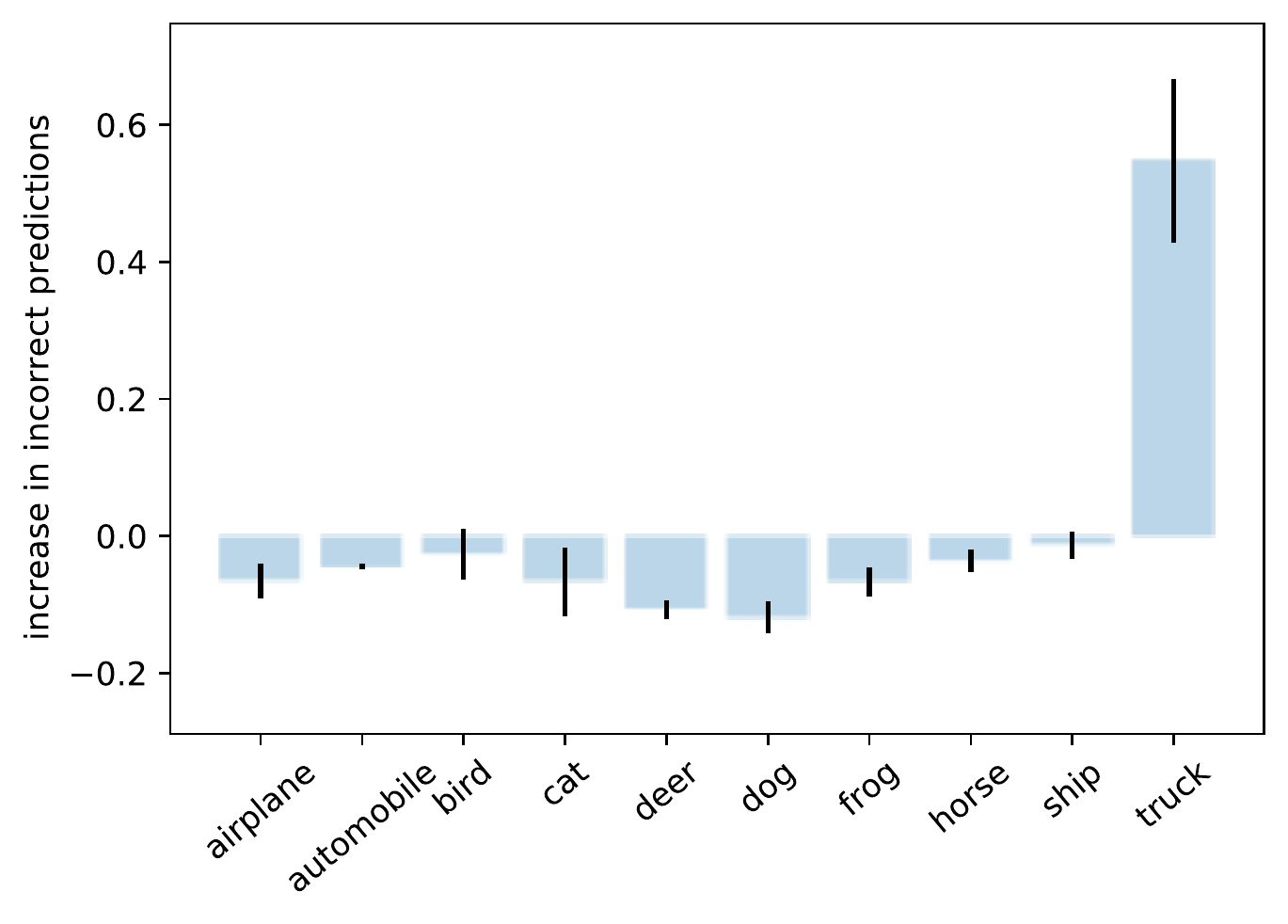}
    \end{minipage}
    \hfill 
    \begin{minipage}[t]{0.49\linewidth} 
        \centering
        \includegraphics[width=0.75\linewidth]{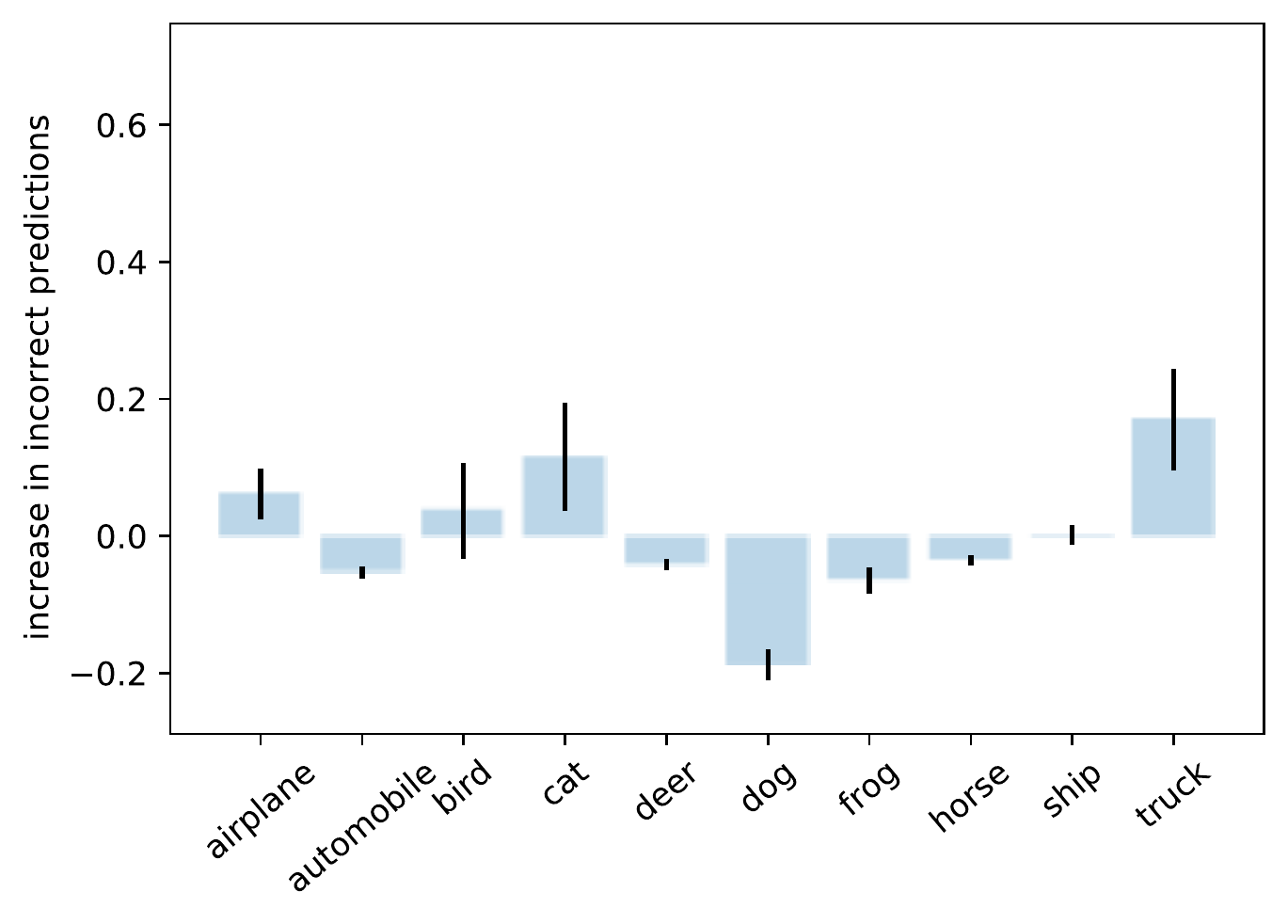}
    \end{minipage}
    \newline
    \begin{minipage}[t]{0.49\linewidth} 
        \centering
        \includegraphics[width=0.75\linewidth]{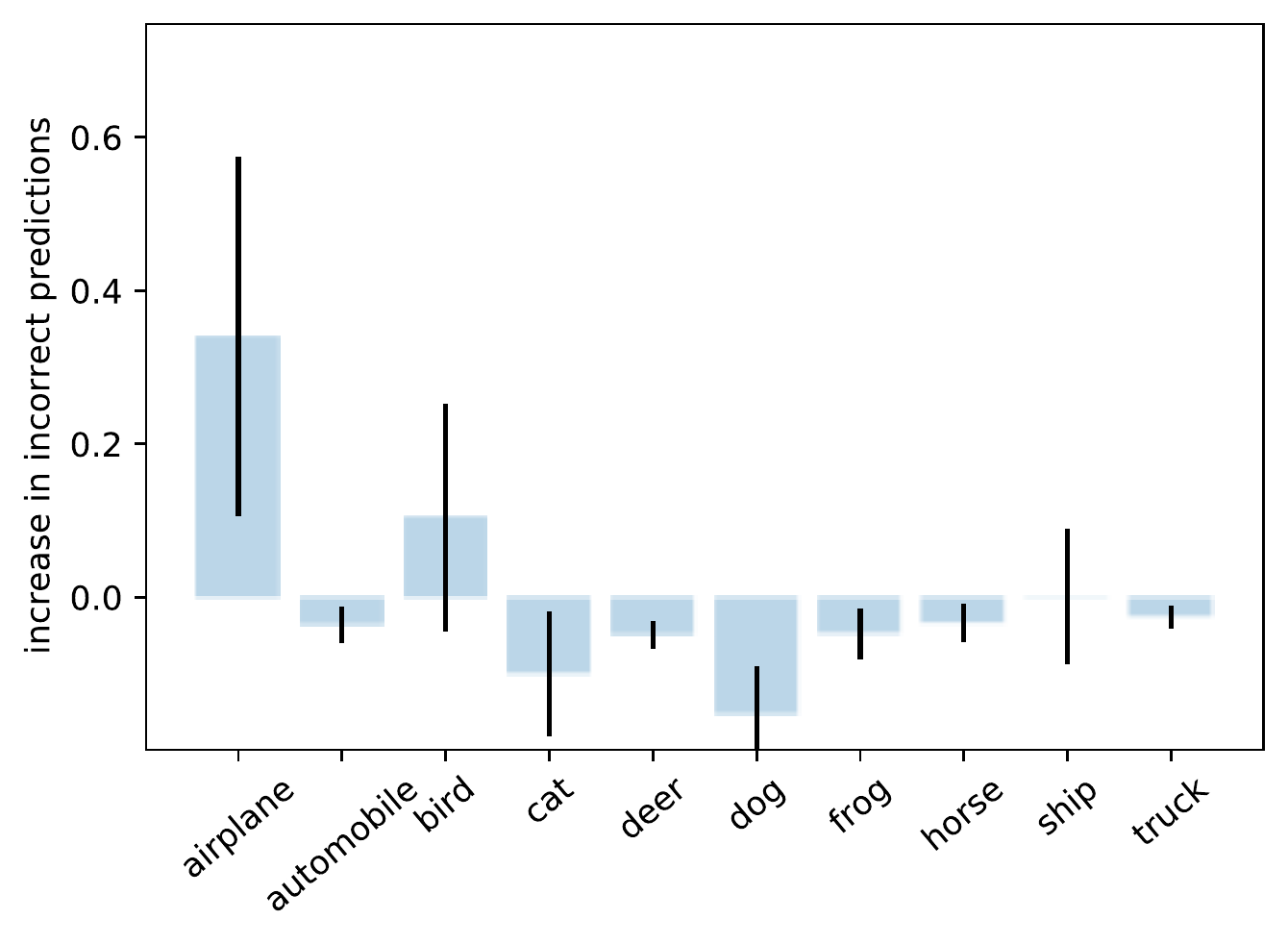}
    \end{minipage}
    \hfill 
    \begin{minipage}[t]{0.49\linewidth} 
        \centering
        \includegraphics[width=0.75\linewidth]{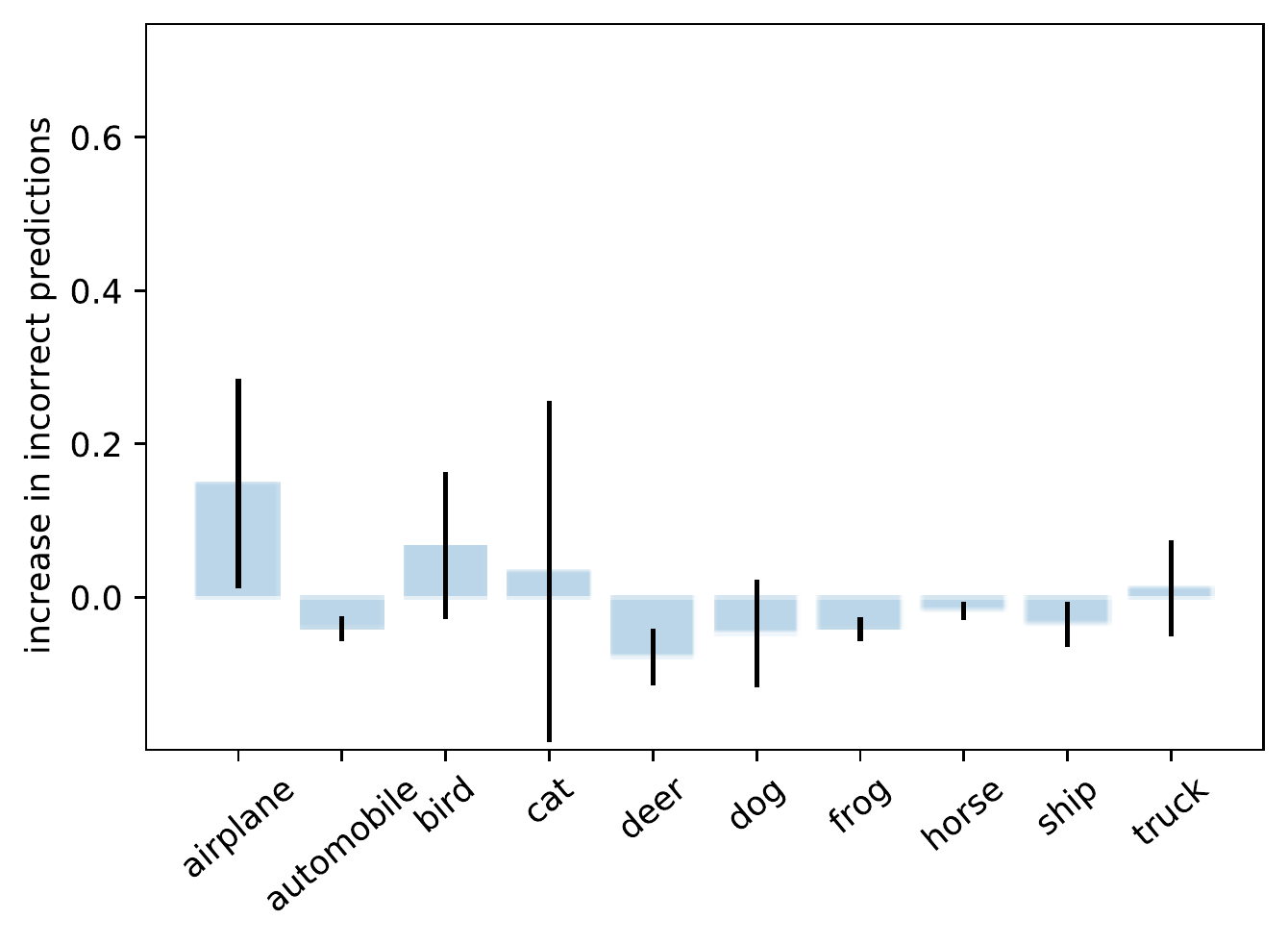}
    \end{minipage}
    \caption{Difference between wrongly predicted classes when testing on original data versus \cutout{} images. Left to right, top to bottom: basic, \mixup, \cutmix, \fmix. The evaluated models from left to right, top to bottom are trained on \cifar{10} with: no mixed-data augmentation (basic), \mixup, \cutmix, and \fmix.}
    \label{fig:wrong_pred_cutout_c10}
\end{figure*}

\begin{figure*}[!h]
    \begin{minipage}[t]{0.49\linewidth}
        \centering
        \includegraphics[width=0.75\linewidth]{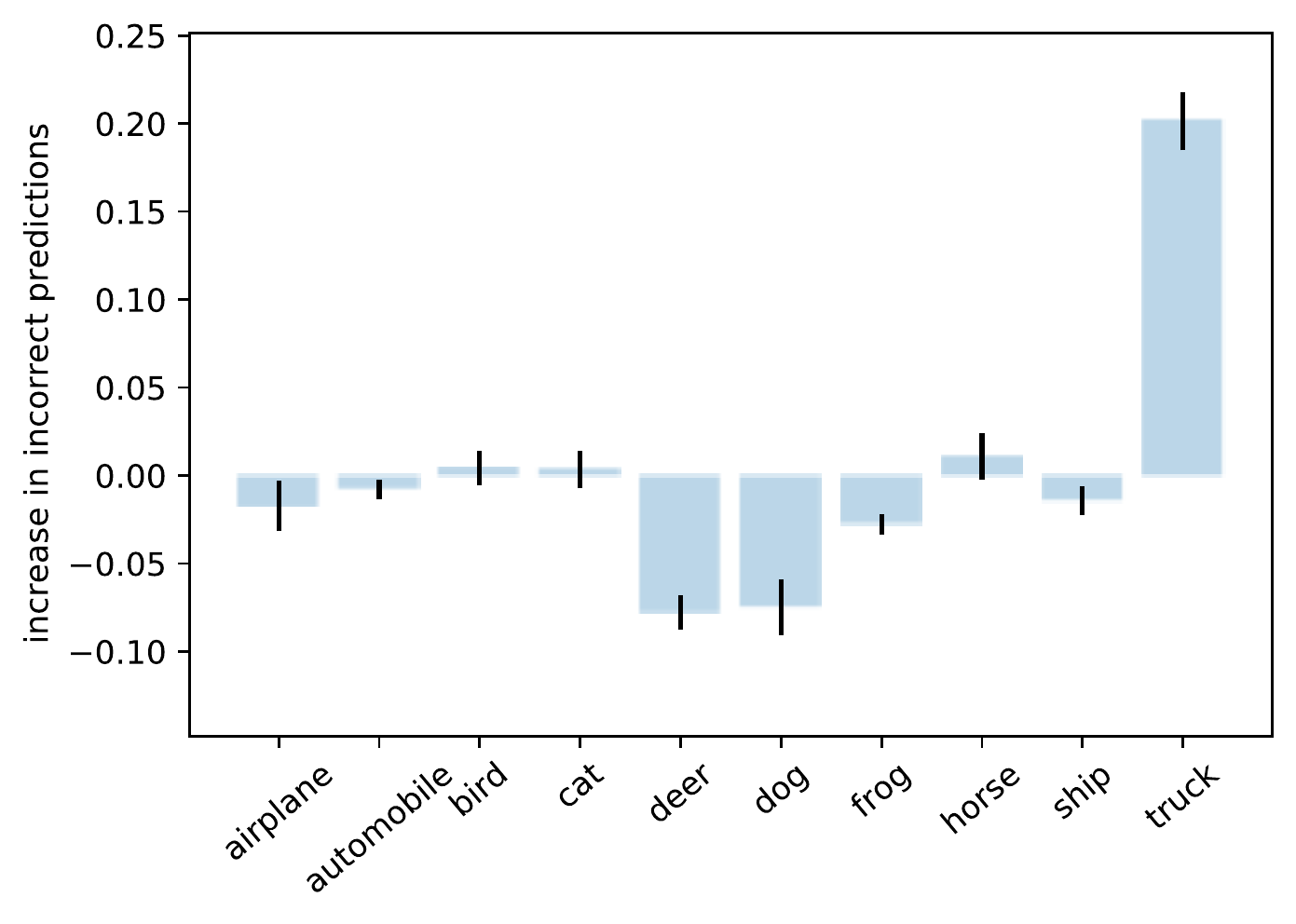}
    \end{minipage}
    \hfill 
    \begin{minipage}[t]{0.49\linewidth} 
        \centering
        \includegraphics[width=0.75\linewidth]{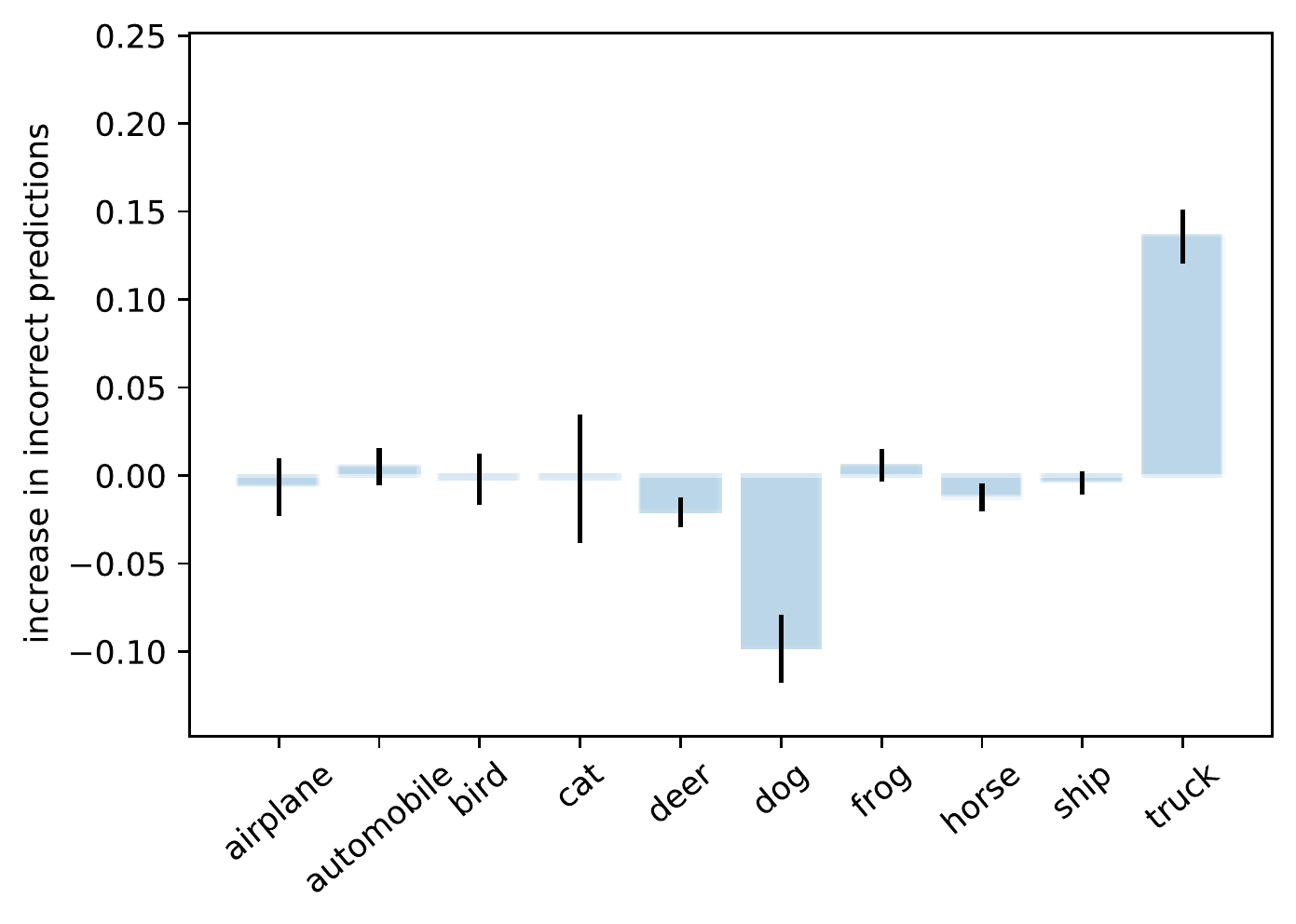}
    \end{minipage}
    \newline
    \begin{minipage}[t]{0.49\linewidth} 
        \centering
        \includegraphics[width=0.75\linewidth]{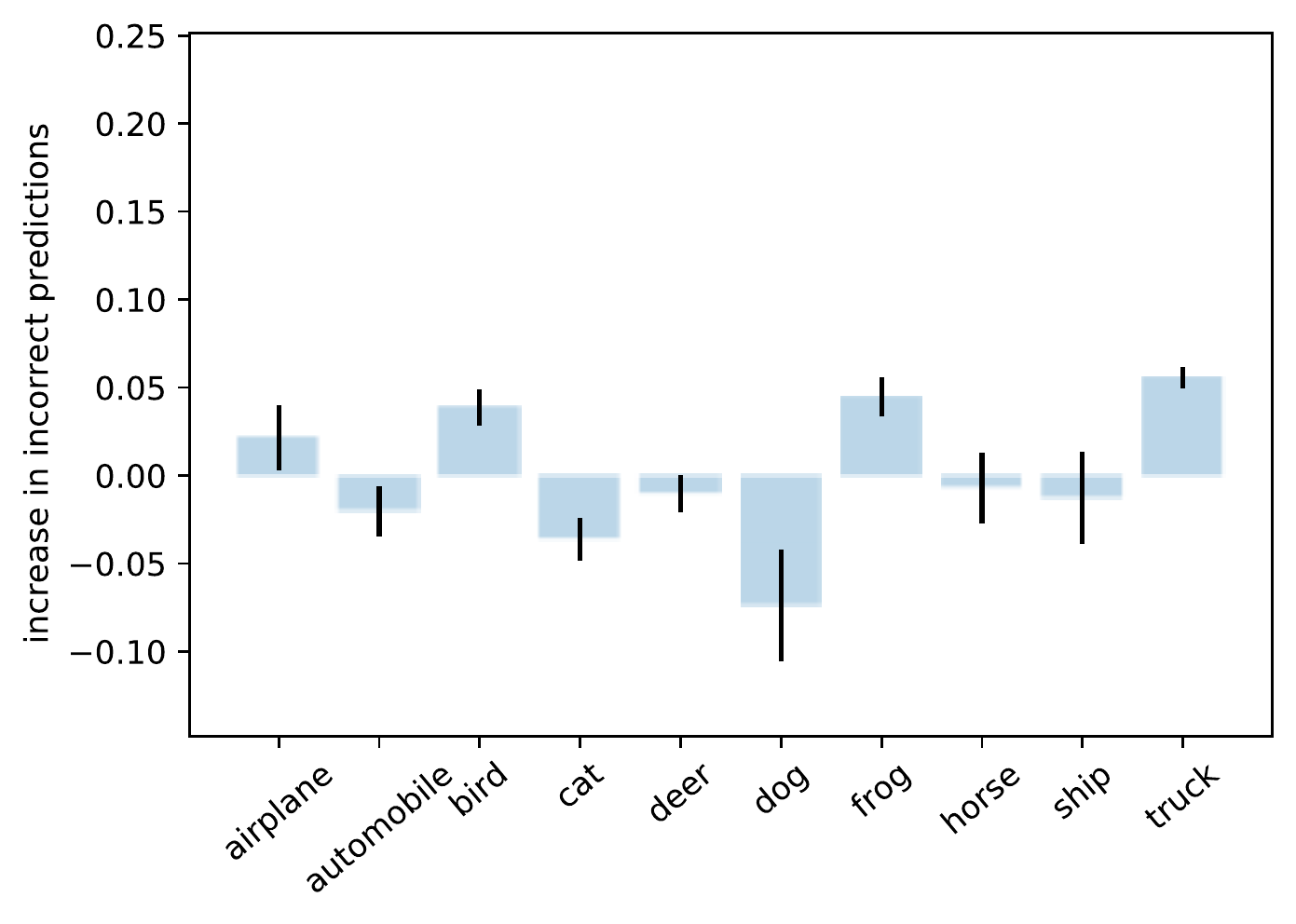}
    \end{minipage}
    \hfill 
    \begin{minipage}[t]{0.49\linewidth} 
        \centering
        \includegraphics[width=0.75\linewidth]{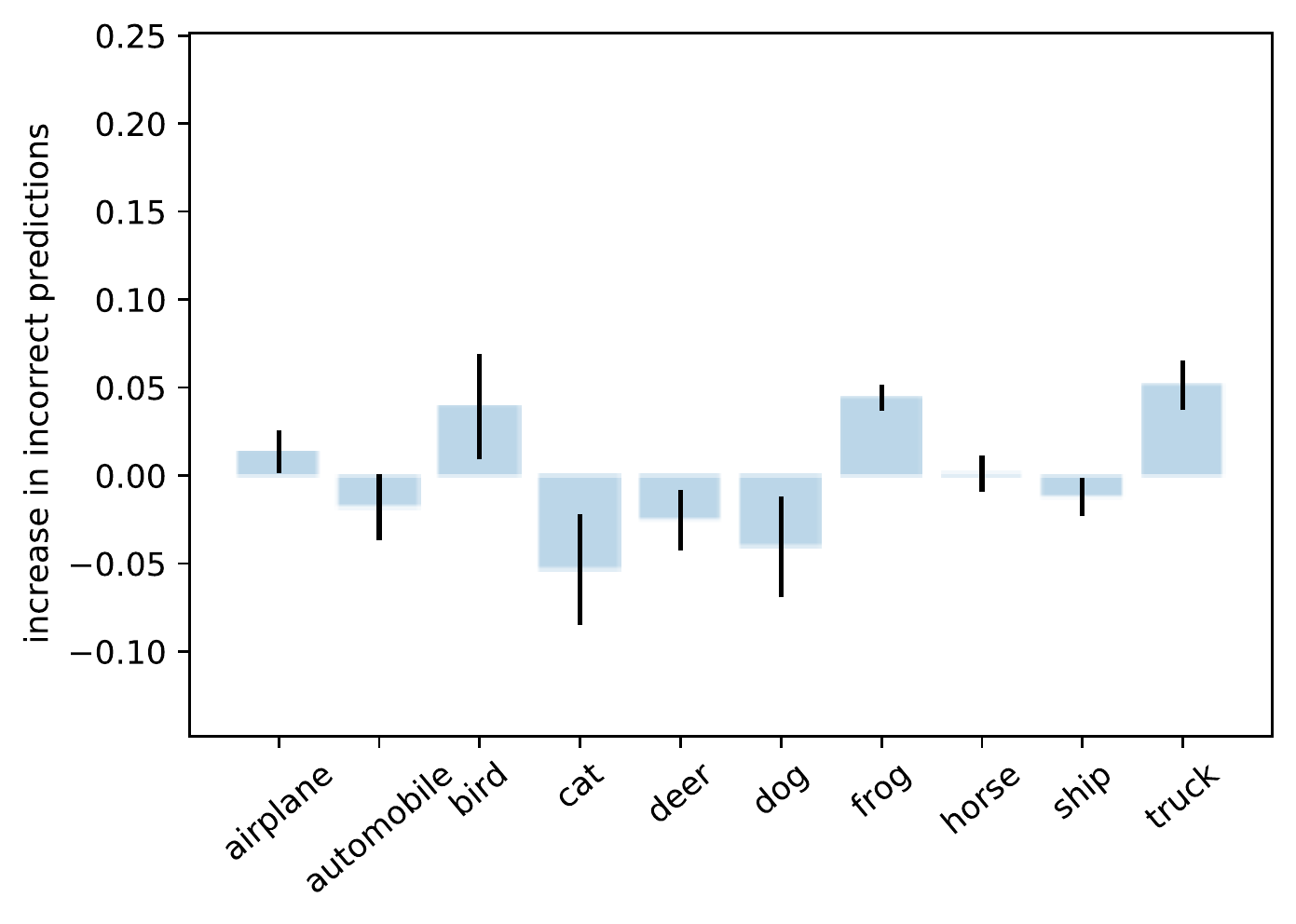}
    \end{minipage}
    \caption{Difference between wrongly predicted classes when testing on original data versus \cutmix images. Left to right, top to bottom: basic, \mixup, \cutmix, \fmix. The evaluated models from left to right, top to bottom are trained on \cifar{10} with: no mixed-data augmentation (basic), \mixup, \cutmix, and \fmix.}
    \label{fig:wrong_pred_cutmix_c10}
\end{figure*}

In this section we experiment with alternative masking methods when computing \cutocc. We note that the bias exists when occluding with patches taken from images belonging to different data sets (Table~\ref{tab:cutmixing}). 
Figure~\ref{fig:wrong_pred_cutmix_c10} gives a visual account of the results obtained for \cifar{10} when mix-patching, while the results obtained when occluding with a uniform patch are given in Figure~\ref{fig:wrong_pred_cutout_c10}.
Note that when mix-patching, for Fashion~MNIST we use MNIST, for Tiny~\imagenet we use \imagenet, while for \cifar{10} we mix with \cifar{100} and vice versa. Since \imagenet images are significantly larger than those of the other data sets, mixing would imply padding large areas, which would give results very similar to uniform patching.
We also experiment with VGG models, where on \cifar{10} the basic has a DI index of $0.80_{\pm 0.40} $ compared to  $0.18_{\pm 0.11} $ of \mixup.

\begin{table}[!h]
        \centering
    \caption{DI index for occluding with images from another data set.} \label{tab:cutmixing}
    \begin{tabulary}{\linewidth}{lLLLL}
    \toprule
            & basic  & \mixup & \fmix   & \cutmix  \\
    \midrule
    \cifar{10} & $0.18_{\pm 0.05}$ &$0.39_{\pm 0.15}$ &$0.12_{\pm 0.11}$ &$0.08_{\pm 0.06}$\\
    \cifar{100} & $0.48_{\pm 0.09}$& $0.61_{\pm 0.27}$& $0.90_{\pm 0.15}$& $1.25_{\pm 0.25}$\\
    Fashion & $3.40_{\pm 0.29}$ & $3.06_{\pm 1.07}$ & $1.81_{\pm 0.55}$ & $2.61_{\pm 0.80}$\\
    Tiny& $0.25_{\pm 0.12}$& $0.17_{\pm 0.04}$& $0.06_{\pm 0.03}$& $0.12_{\pm 0.04}$\\
    \bottomrule
    \end{tabulary}
\end{table}
\newpage

\newpage
We then use masks sampled from Fourier space (Table~\ref{tab:fout}) and note that even for these irregularly shaped distortions, we can identify a gap in most cases. The only exception is in the case of Fashion MNIST. It must be stressed that although all the models we experimented with presented Data Interference for this problem, this does not exclude the possibility of constructing a different model that is insensitive to this distortion. 
For example, we identify a gap for this problem when mix-masking (DI index of $4.09_{\pm 1.74}$ for the basic model as opposed to $1.87_{\pm 0.27}$ for a model trained on images that were masked out using \fmix-like masks).
Thus, when occluding with a particular shape we implicitly disfavour models in which learnt representations are related to the features introduced by that shape.

\begin{table}[h]
        \centering
    \caption{DI index for patching using masks sampled from Fourier space.} \label{tab:fout}
    \vskip 0.15in
    \begin{tabulary}{\linewidth}{lLLLL}
    \toprule
            & basic  & \mixup & \fmix   & \cutmix  \\
    \midrule
    \cifar{10} & $2.08_{\pm 1.13}$& $1.79_{\pm 1.09}$& $1.32_{\pm 0.99}$& $4.21_{\pm 1.23}$\\
    \cifar{100} & $4.06_{\pm01.47}$ & $3.11_{\pm02.29}$ & $9.90_{\pm14.32}$ & $2.89_{\pm05.36}$\\
    Fashion & $49.55_{\pm 20.4}$& $40.69_{\pm 21.6}$& $27.87_{\pm 17.5}$& $61.04_{\pm 17.9}$\\
    Tiny & $4.37_{\pm 0.85}$ &$6.95_{\pm 1.84}$ &$3.60_{\pm 1.73}$ &$5.92_{\pm 4.38}$\\
    \imagenet & $3.27$ & $2.24$ & $6.08$ & $-$\\
    \bottomrule
    \end{tabulary}
\end{table}

\begin{figure*}[h] 
    \begin{minipage}{0.49\linewidth}
        \centering
        \includegraphics[width=0.9\linewidth]{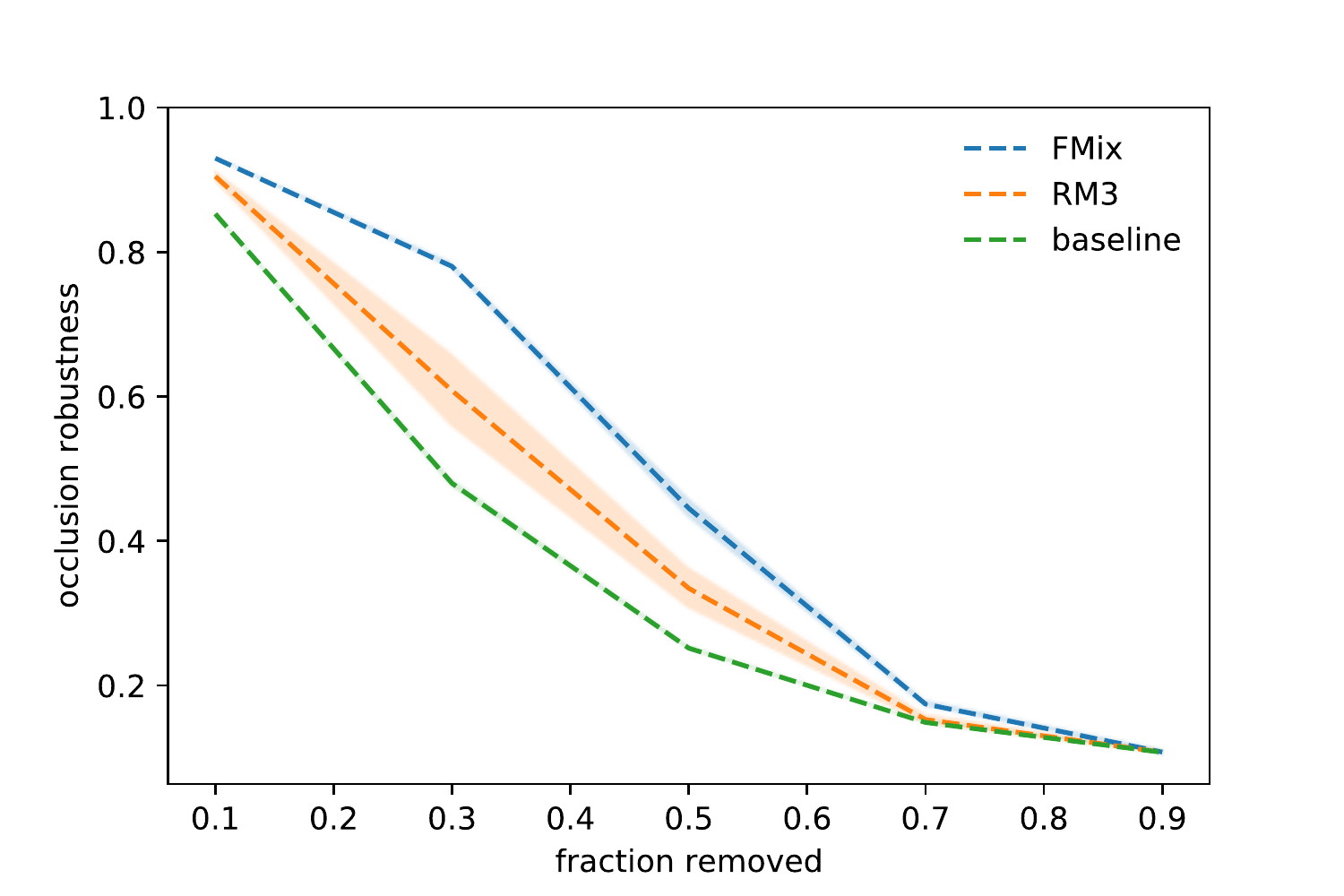}
    \end{minipage}
    \hfill 
    \begin{minipage}{0.49\linewidth}
        \centering
        \includegraphics[width=0.9\linewidth]{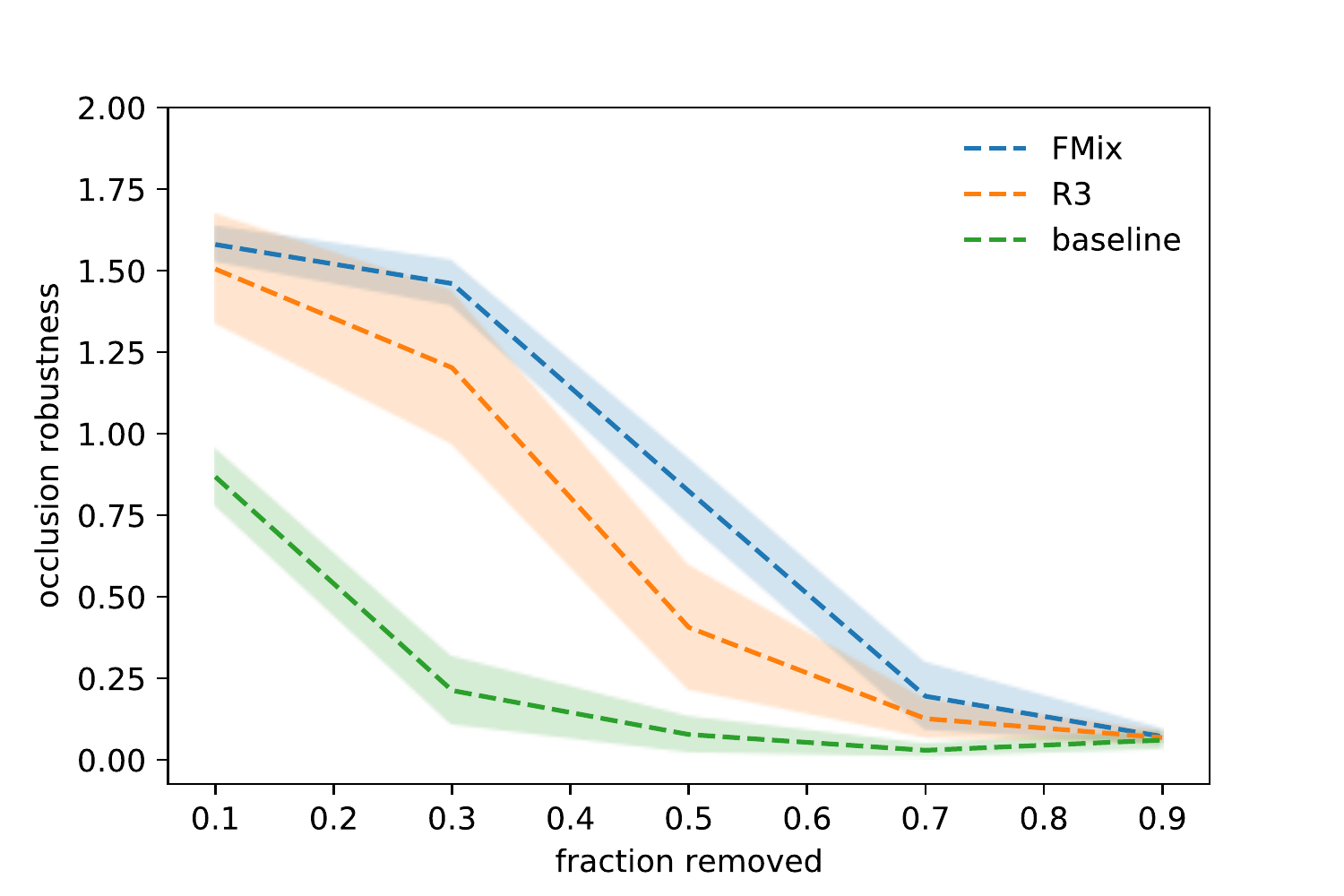}
    \end{minipage}
    \caption{ \cutocc (left) and \iocc (right). Note that there is a difference in scale and the two should not be directly compared. We are rather interested in how the methods situate the different augmentation with respect to each other. It is important to notice that when measuring the robustness with \cutocc, RM3 appears significantly less robust than \fmix due to its sensitivity to patching with rectangles. On the other hand, \iocc highlights the robustness specific to \fmix.
    }
    \label{fig:3msks}
\end{figure*}

Figure~\ref{fig:3msks} also gives the results for \cutocc and \iocc for training with 3 random masks sampled from Fourier space. 

\subsection{BagNet Shape and Texture Accuracy} \label{sup:tiny_jig}
\begin{figure}[h]
    \centering
    \includegraphics[width=0.5\linewidth]{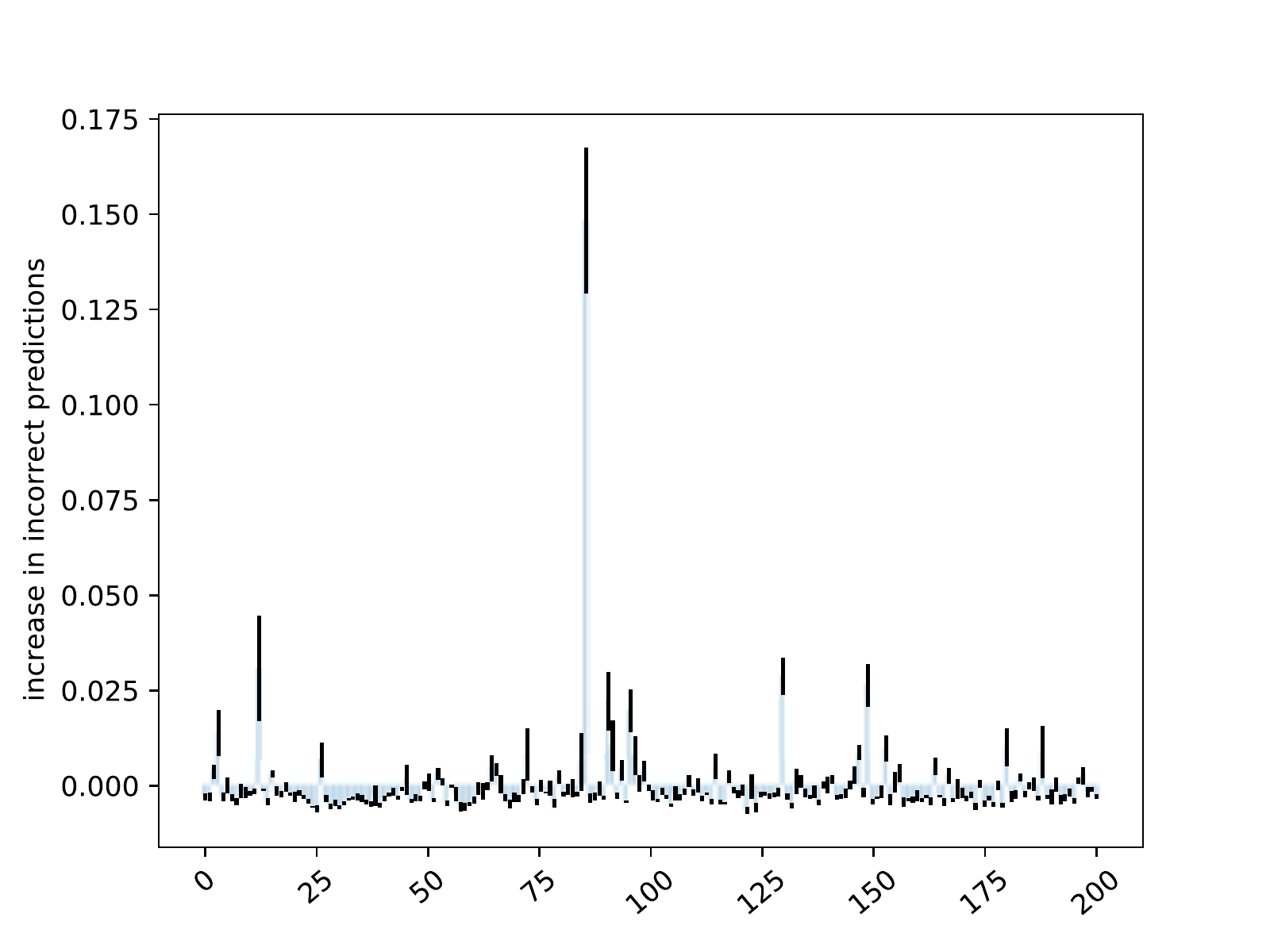}
    \vskip -0.05in
    \caption{Difference between wrongly predicted classes when testing on original Tiny \imagenet data versus \cutout{} images.} 
    \label{fig:bag9_bias}
\end{figure}

We evaluate on the GST data set BagNet9 models trained on Tiny~\imagenet and present the results in Table~\ref{sup_tab:tiny_bag}. Despite the basic model displaying a bias towards predicting one of the classes when presented with patch-shuffled images (see Figure~\ref{fig:bag9_bias}), once again it does not have a lower texture or higher shape bias than the masked-based augmentations.

\begin{table}[!h]
    \centering
    \caption{Shape and texture accuracy of BagNet9 models on the GST data set.}
    \label{sup_tab:tiny_bag}
    \vskip 0.15in
    \begin{tabulary}{\linewidth}{lLL}
    \toprule
                & Shape             & Texture \\
    \midrule
    basic    & $11.29_{\pm 0.15}$ & $18.90_{\pm0.66}$ \\
    \mixup      & $11.04_{\pm 0.29}$ & $12.56_{\pm 1.26}$\\ 
    \fmix       & $11.06_{\pm 0.48}$ & $17.47_{\pm1.74}$\\
    \cutmix     & $10.76_{\pm 0.27}$ & $20.28_{\pm 0.88}$\\
    \bottomrule
    \end{tabulary}
    \vskip -0.1in
\end{table}

\subsection{Data Interference across Architectures}
\label{sup:architectures}

To verify if a shape bias evaluation based on patch-shuffling would give unfair results when comparing across architectures, we compute the \di index for a number of models trained with the basic approach (without mixed data augmentation) on \cifar{10}.
The DI index of models such as WideResNet and ResNet is high ($0.95_{\pm0.20}$ and $1.27_{\pm0.39}$ respectively), while for PyramidNet, BagNet17 and BagNet9 it is small ($0.26$, $0.53_{\pm 13}$, $54_{\pm0.16}$). 
DenseNet ($0.66_{\pm 0.48}$) and VGG ($0.60_{\pm 0.15}$) have comparable DI indices.  

Thus, we find that intensity with which distortions interfere with learnt representations is different for different architectures. 
Comparing robustness to occlusion using \cutocc would give biased results when comparing architectures trained under the same conditions.

\section{Further Results on \iocc Experiments} \label{sup:further_results_iocc}

\subsection{Alternative \cutocc} \label{sup:unifrom}

Table~\ref{tab:unif} gives the DI index when forcing the occluding patch to lie within image boundaries for patch sizes sampled uniformly from [0.1,\, 1]. We also experimented with sampling occluder sizes from a Beta(2,1) distribution and obtained similar results.
Naturally, the \di index is more pronounced and we once again notice a gap for all data sets.

\begin{table}[h]
        \centering
    \caption{DI index for sampling occluder size from a uniform distribution when the patch is restricted to lying within image boundaries and the size is sampled from [0.1, \,1] uniformly.} \label{tab:unif}
    \vskip 0.15in
    \begin{tabulary}{\linewidth}{lLLLL}
    \toprule
         & basic  & \mixup & \fmix   & \cutmix  \\
    \midrule
    \cifar{10}& $5.71_{\pm 0.18}$ & $0.51_{\pm 0.41}$ & $0.77_{\pm 0.75}$ & 
$2.88_{\pm 1.31}$ \\
    \cifar{100} & $27.81_{\pm 5.23}$ & $4.40_{\pm 3.25}$ & $4.12_{\pm 3.12}$ & $5.13_{\pm 6.68}$ \\
    Fashion & $0.93_{\pm0.98}$ & $3.46_{\pm0.58}$ & $0.99_{\pm0.99}$ & $0.69_{\pm0.93}$ \\
    Tiny & $13.72_{\pm 2.46}$ & $5.47_{\pm 1.48}$ & 
$4.10_{\pm 2.69}$ & $6.31_{\pm 6.76}$ \\
    \imagenet &$0.25$ & $0.48$ & $0.14$ &  $-$\\

    \bottomrule
    \end{tabulary}
    \vskip -0.1in
\end{table}

\subsection{Occluding with Images from Another Data Set} \label{sup:mix_nomix}

\begin{figure*}[!h]
\vskip 0.2in
    \begin{minipage}{0.49\linewidth}
        \centering
        \includegraphics[width=0.9\linewidth]{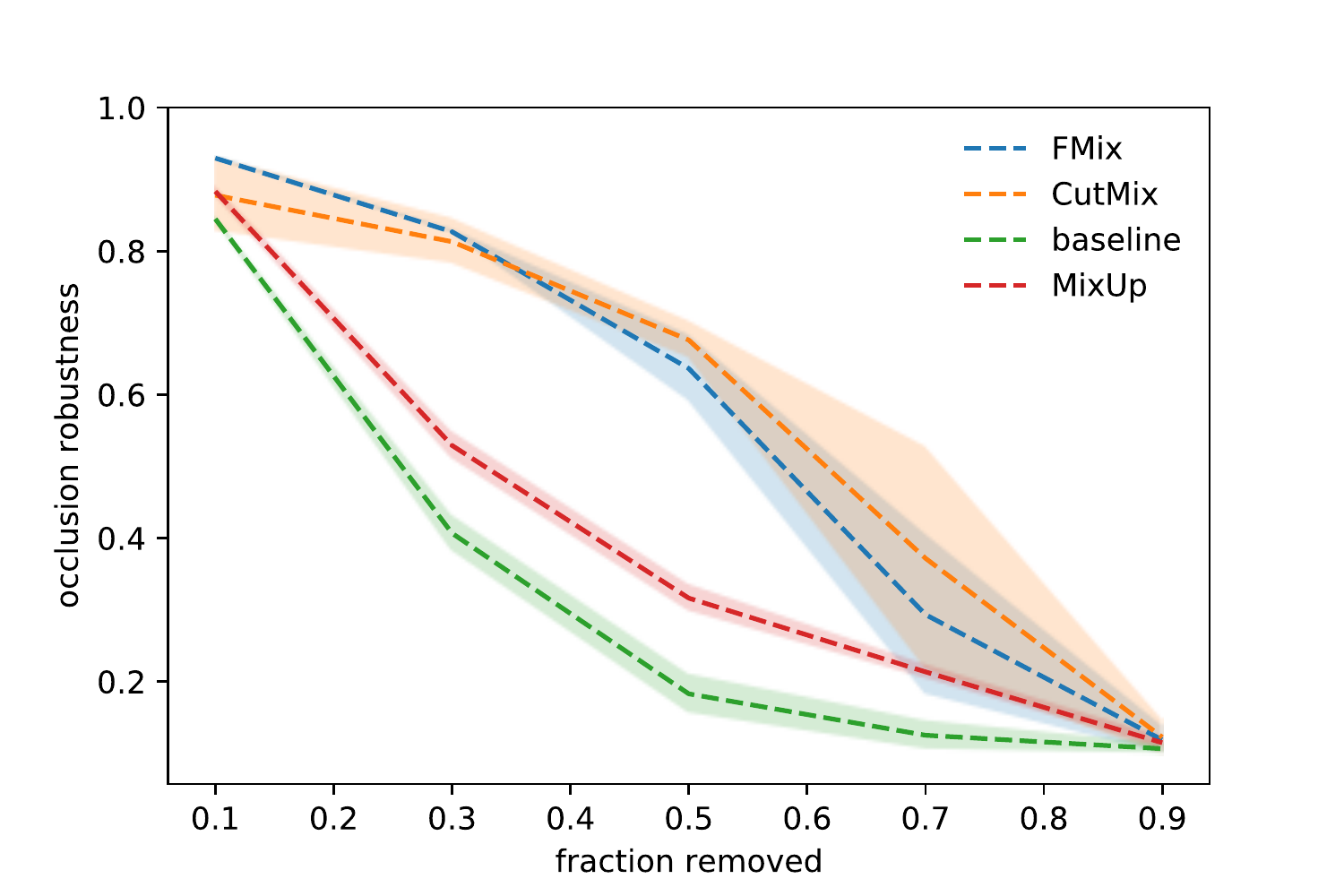}
        
    \end{minipage}
    \hfill 
    \begin{minipage}{0.49\linewidth} 
        \centering
        \includegraphics[width=0.9\linewidth]{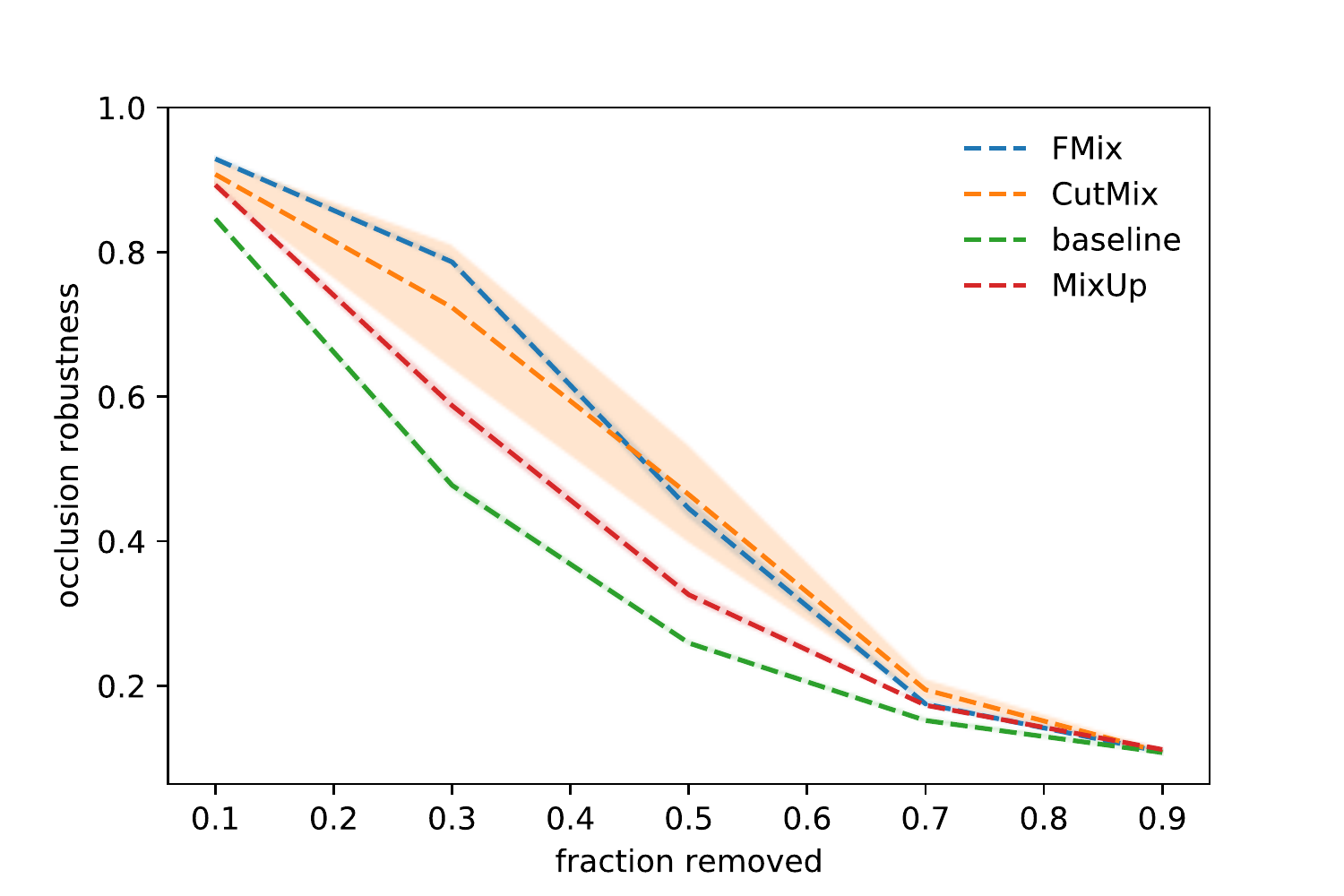}
        
    \end{minipage}
    \newline
    \begin{minipage}{0.49\linewidth} 
        \centering
        \includegraphics[width=0.9\linewidth]{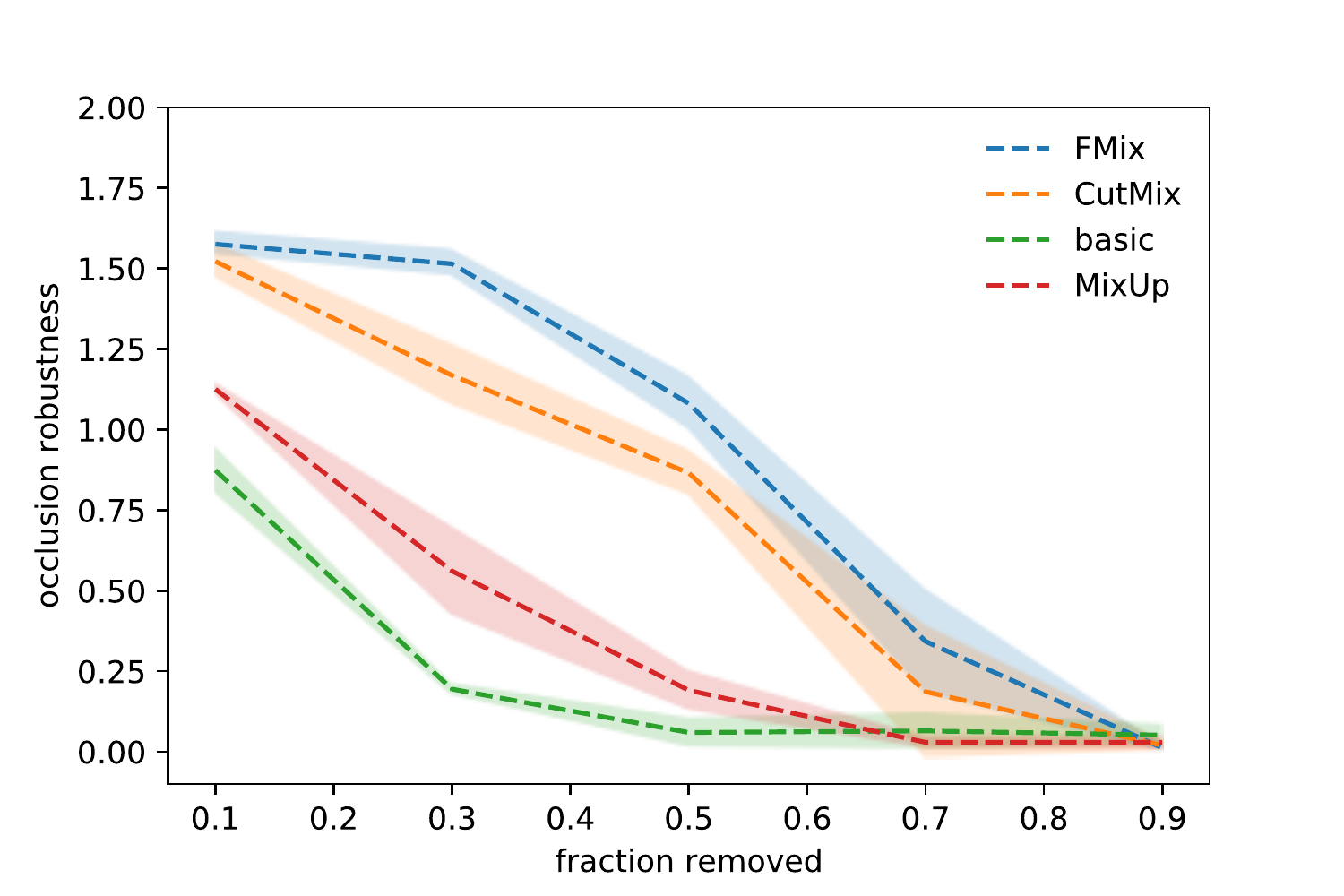}
        
    \end{minipage}
    \hfill 
    \begin{minipage}{0.49\linewidth} 
        \centering
        \includegraphics[width=0.9\linewidth]{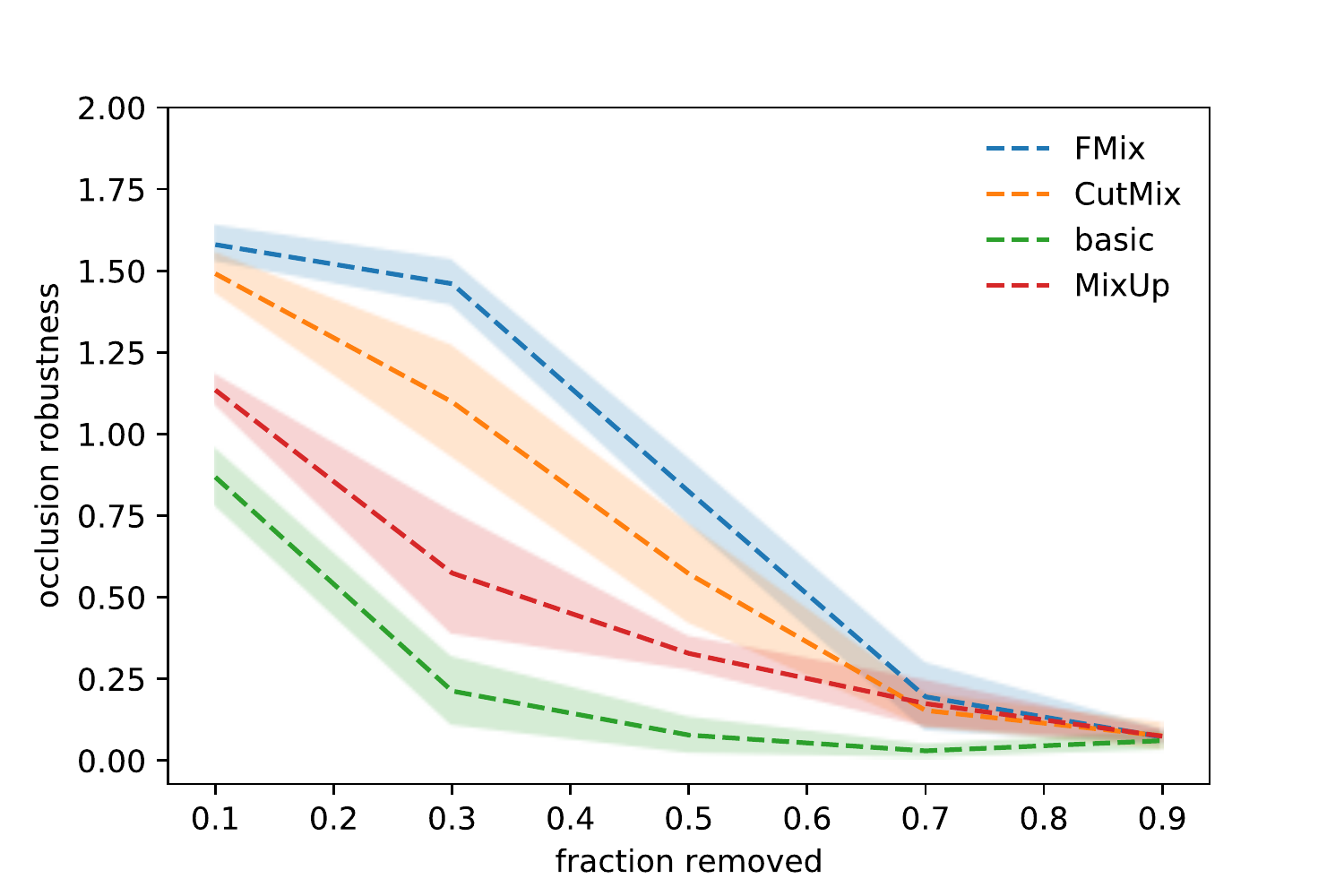}
         
    \end{minipage}
    \caption{Comparison of metric sensitivity to textured occlusion. 
    \cutocc (top) versus \iocc (bottom) when occluding with uniform patches (left hand-side) and non-uniform patches (right hand-side).
    Uniform occlusion refers to superimposing uniform patches over \cifar{10} images, while nonuniform refers to superimposing part of \cifar{100} samples. Nonuniform \cutocc provides significantly different results to its regular counterpart, whereas \iocc provides more consistent results.}
    \label{fig:mix_nomix}
    \vskip -0.2in
\end{figure*}

Since \cutocc does not account for the bias introduced by the occluding method, it is expected that changing the patch to a non-uniform one would greatly affect the results. For \cifar{10} models, Figure~\ref{fig:mix_nomix} presents the results of occluding with uniform patches versus patches extracted from \cifar{100} images.
\iocc better rules out the specifics of the occluding patch, its uniform version giving similar results to the non-uniform one, whereas \cutocc pushes everything together.

\subsection{Randomising Labels} \label{sup:rand_lbls}

\begin{figure}[h]
\vskip 0.2in
    \centering
        \includegraphics[width=0.6\linewidth]{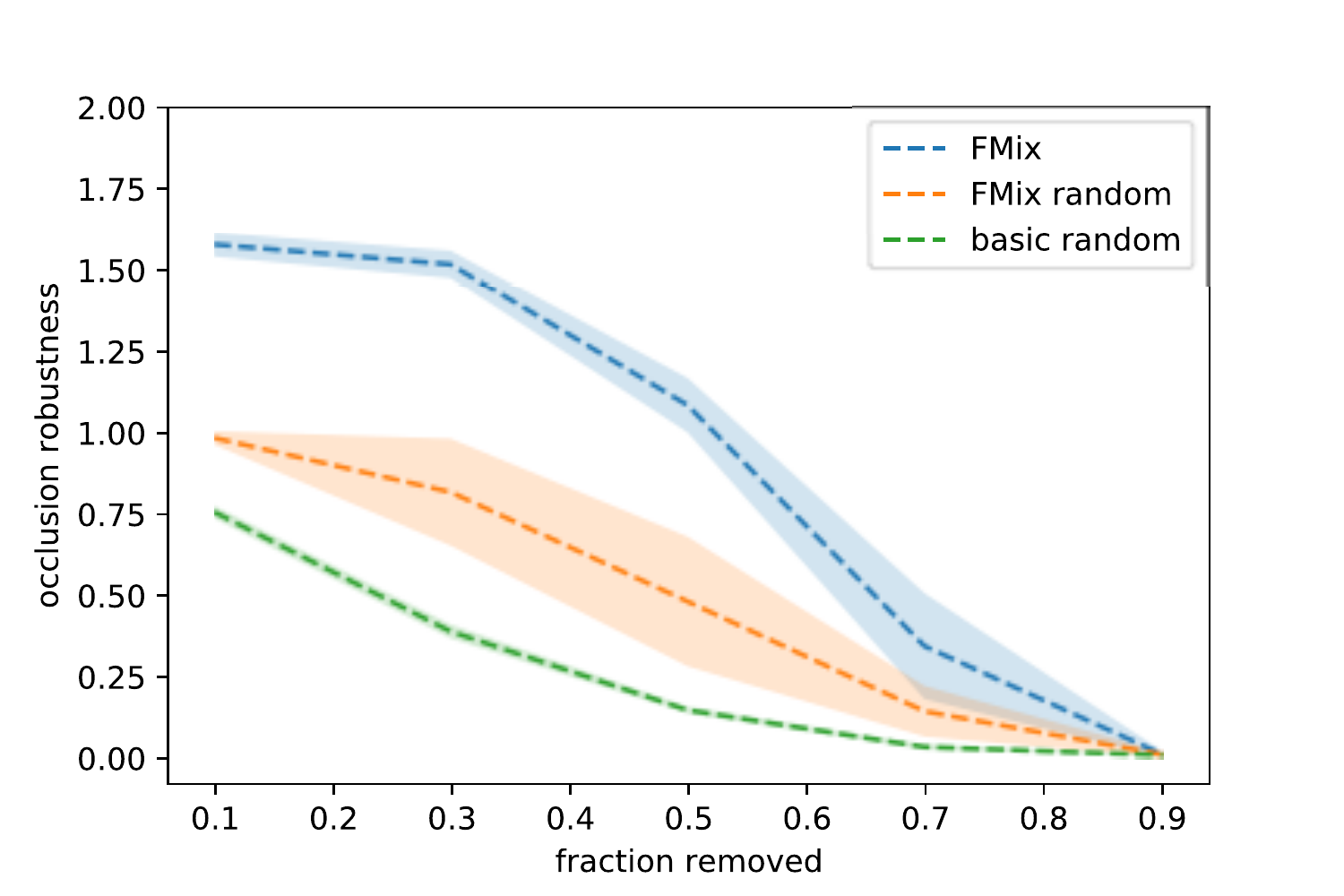}
         \caption{\iocc results for training with clean and corrupted labels for basic and \fmix augmentation.} 
        \label{fig:rand}
        \vskip -0.2in
\end{figure}

To assess the sensitivity of \cutocc and \iocc to the overall performance of the model, we also experiment with randomising all the labels of the \cifar{10} data set.
When evaluated on the unaugmented training data, all the basic models achieve 100\% accuracy, while the \fmix models reach $99.99_{\pm0.01}$. 
Since all labels are corrupted, the accuracy on the test set before and after occlusion is no greater than random. 
However, the robustness of the augmentation-trained model can be seen on the training data, as captured by our metric (see Figure~\ref{fig:rand}).
On the other hand, \cutocc makes no distinction between learning with regular and augmented data (Table~\ref{tab:rand_oriz}).
Despite being such a peculiar case, it shows the comprehensiveness gained by accounting for the degradation on test data in relation to that on train.

\begin{table}[h]
    \centering
    \caption{Robustness to occluding with patches covering $50\%$ of each image. The models are trained with and without masking augmentation on data with randomised labels. 
    \cutocc makes no difference between regular and augmented training.} \label{tab:rand_oriz}
    \vskip 0.15in
    \begin{tabulary}{\linewidth}{lLLL}
    \toprule
     & basic random & \fmix random & \fmix clean \\
     \midrule
    \cutocc &$10.24_{\pm0.27}$ &$9.78_{\pm0.18}$ &$63.63_{\pm4.54}$ \\
    \iocc   &$14.63_{\pm1.12}$ & $47.94_{\pm19.84}$ & $82.36_{\pm10.06}$\\
    \bottomrule
    \end{tabulary}
    \vskip -0.1in
\end{table}

\subsection{\iocc Numerator and Denominator}

In this section we give the values of the composing elements of \iocc for models trained on \cifar{10}.
While the numerator depends on the size of the occluder, the denominator is a fixed term: the generalisation gap.
Results are presented in Table~\ref{tab:num-denum}.

\begin{table*}[h]
    \centering
    \caption{Value of the numerator (accuracy gap when occluding a proportion of the image) and denominator (generalisation gap) for models trained on \cifar{10}.} \label{tab:num-denum}
    \vskip 0.15in
    \begin{tabulary}{\linewidth}{lLLLLL|L}
    \toprule
    &  \multicolumn{5}{c}{numerator} & denominator \\
    \midrule
    proportion & 10\% & 30\% & 50\% & 70\% & 90\% & \\
    \midrule
    basic & $4.66_{\pm 0.38}$ & $1.06_{\pm 0.10}$ & $0.32_{\pm 0.24}$ & $0.35_{\pm 0.30}$ & $0.29_{\pm 0.19}$ & $5.47_{\pm 0.10}$\\
    \cutmix & $6.96_{\pm 0.24}$ & $ 5.49_{\pm 0.44}$ & $ 3.78_{\pm 0.30}$ & $ 0.86_{\pm 0.94}$ & $ 0.09_{\pm 0.07}$ & $4.60_{\pm 0.12}$\\
    \fmix & $7.04_{\pm 0.16}$ & $ 7.32_{\pm 0.20}$ & $ 5.07_{\pm 0.39}$ & $ 1.51_{\pm 0.71}$ & $ 0.05_{\pm 0.04}$ & $4.58_{\pm 0.15}$\\ 
    \mixup & $5.43_{\pm 0.09}$ & $ 2.68_{\pm 0.65}$ & $ 0.86_{\pm 0.27}$ & $ 0.13_{\pm 0.08}$ & $ 0.14_{\pm 0.10}$ &  $4.70_{\pm 0.12}$\\ 
    \bottomrule
    \end{tabulary}
    \vskip -0.1in
\end{table*}

\subsection{Approximating \iocc} \label{sup:approximating}

\begin{figure*}[h]
    \begin{minipage}[t]{0.49\linewidth} 
        \centering
        \includegraphics[width=0.85\linewidth]{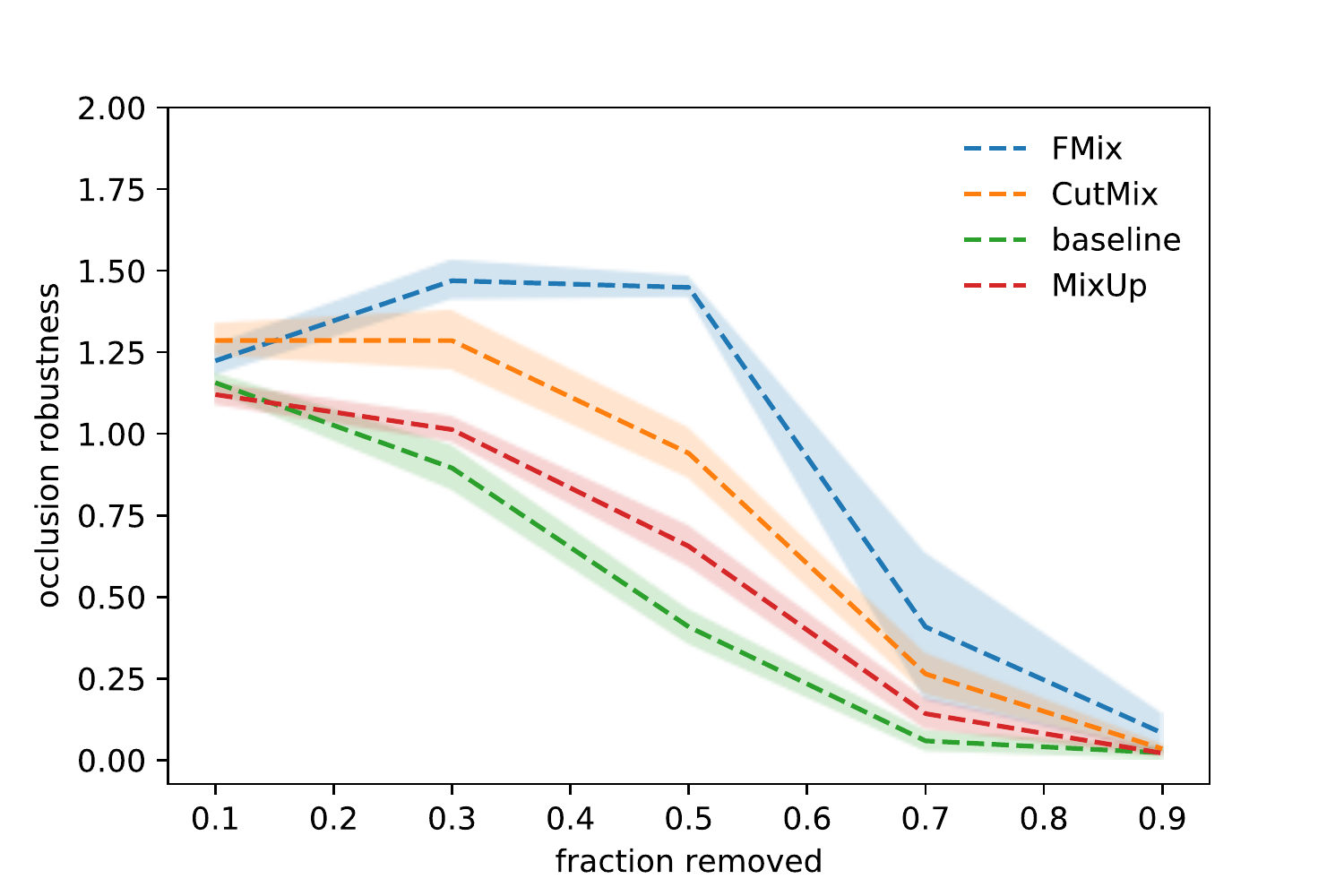}
    \end{minipage}
    \hfill 
    \begin{minipage}[t]{0.49\linewidth} 
        \centering
        \includegraphics[width=0.85\linewidth]{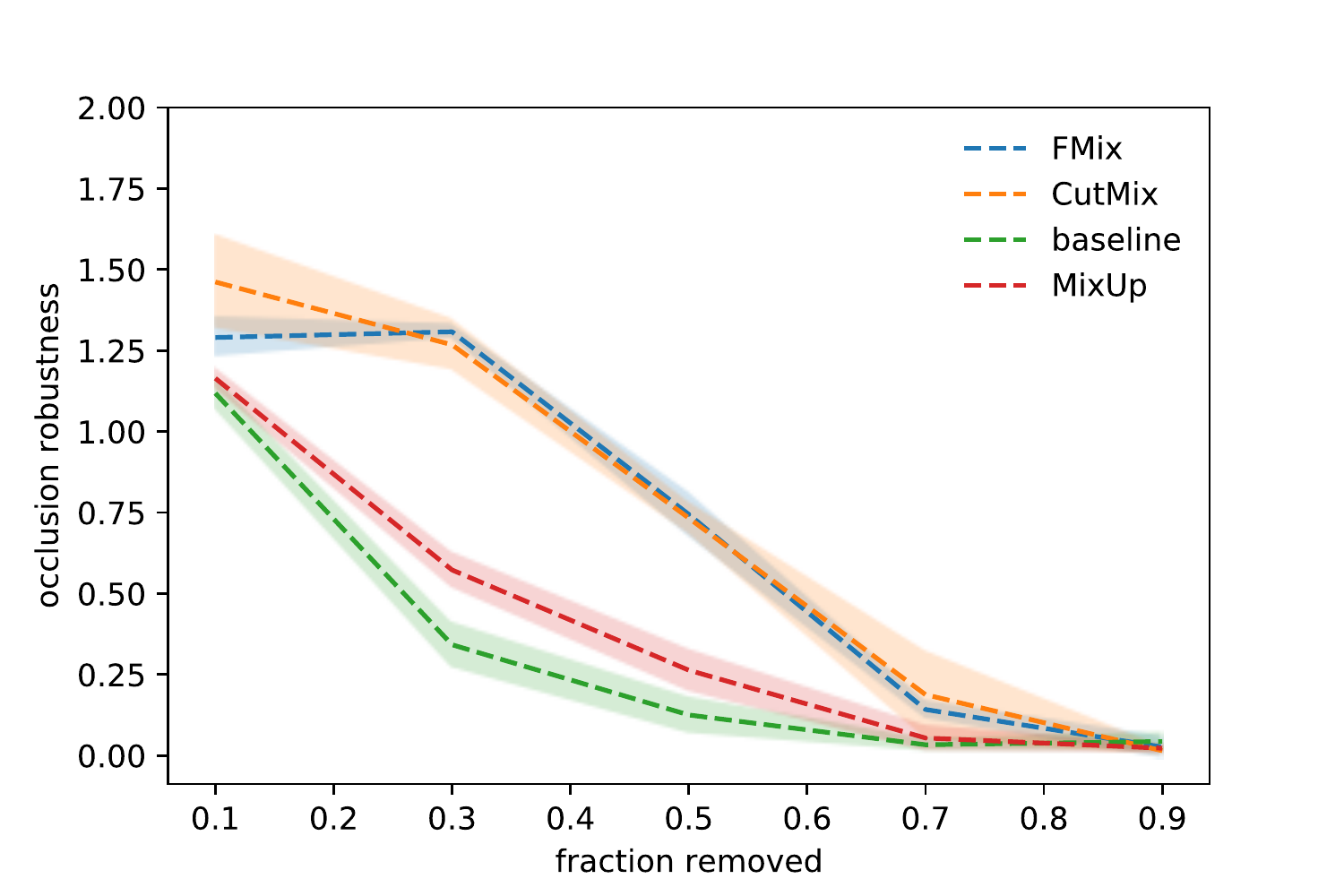}
    \end{minipage}
    \caption{\iocc when using Fourier-sampled masks (left) and rectangular masks (right).Approximating \iocc with random masking can provide a first intuition, but is noisier than using a saliency method.}
    \label{fig:approx}
\end{figure*}

As alternative methods for computing \iocc we experiment both with masks sampled from Fourier space and randomly positioned square patches.
Although using this type of random masking methods for computing $\mathcal{D}^{p}_{train}$ and $\mathcal{D}^{p}_{test}$ in Equation~(1) gives less precise results, it has the advantage of incurring less computation and can be used for rapid model analysis. 
For assessing a model across 5 runs for 6 different levels of occlusion, this method leads to a carbon footprint of 0.05 kgCO$_2$eq as opposed to 1.04 using Grad-CAM.
In Figure~\ref{fig:approx} we present the results obtained with these alternatives. 
We expect both methods to provide overoptimistic results for small patches, while Fourier sampling is expected to give more truthful scores as the size of the patch increases.
On the other hand, the contiguity of \cutout-based occlusion comes at the cost of not determining the robustness to multiple simultaneous occluders. 
This seems to play a role especially in the case of \cutmix augmentation.
Indeed, when superimposing a rectangular patch, it is difficult to differentiate \cutmix from \fmix-trained models.
To confirm that this is caused by the granularity of the occluders and not the shape, we also experiment with occluding using multiple rectangular patches.
We split the images in a $4 \times 4$ grid and occlude i\% of the tiles, obtaining results that are more similar to those obtained when occluding with Fourier-sample patches.
Thus, while significantly noisier, using randomly positioned occluders can provide an alternative for computing \iocc given that one takes into account the number of occluders.

\subsubsection{Removing the Dominant Class} \label{sup:removeDominnat}

\begin{figure*}[h]
\vskip 0.2in
    \begin{minipage}[t]{0.49\linewidth} 
        \centering
        \includegraphics[width=0.85\linewidth]{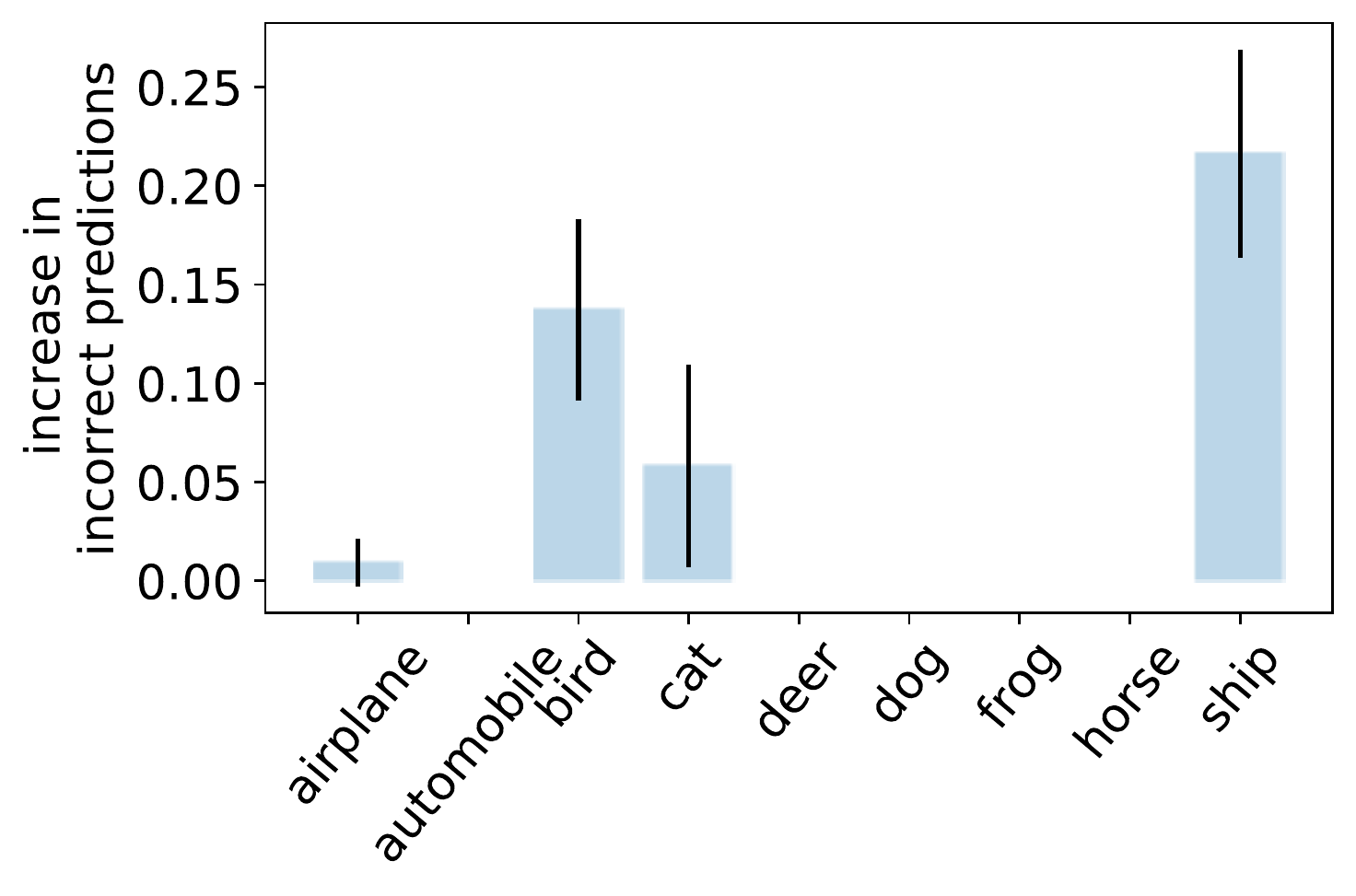}
    \end{minipage}
    \hfill 
    \begin{minipage}[t]{0.49\linewidth} 
        \centering
        \includegraphics[width=0.85\linewidth]{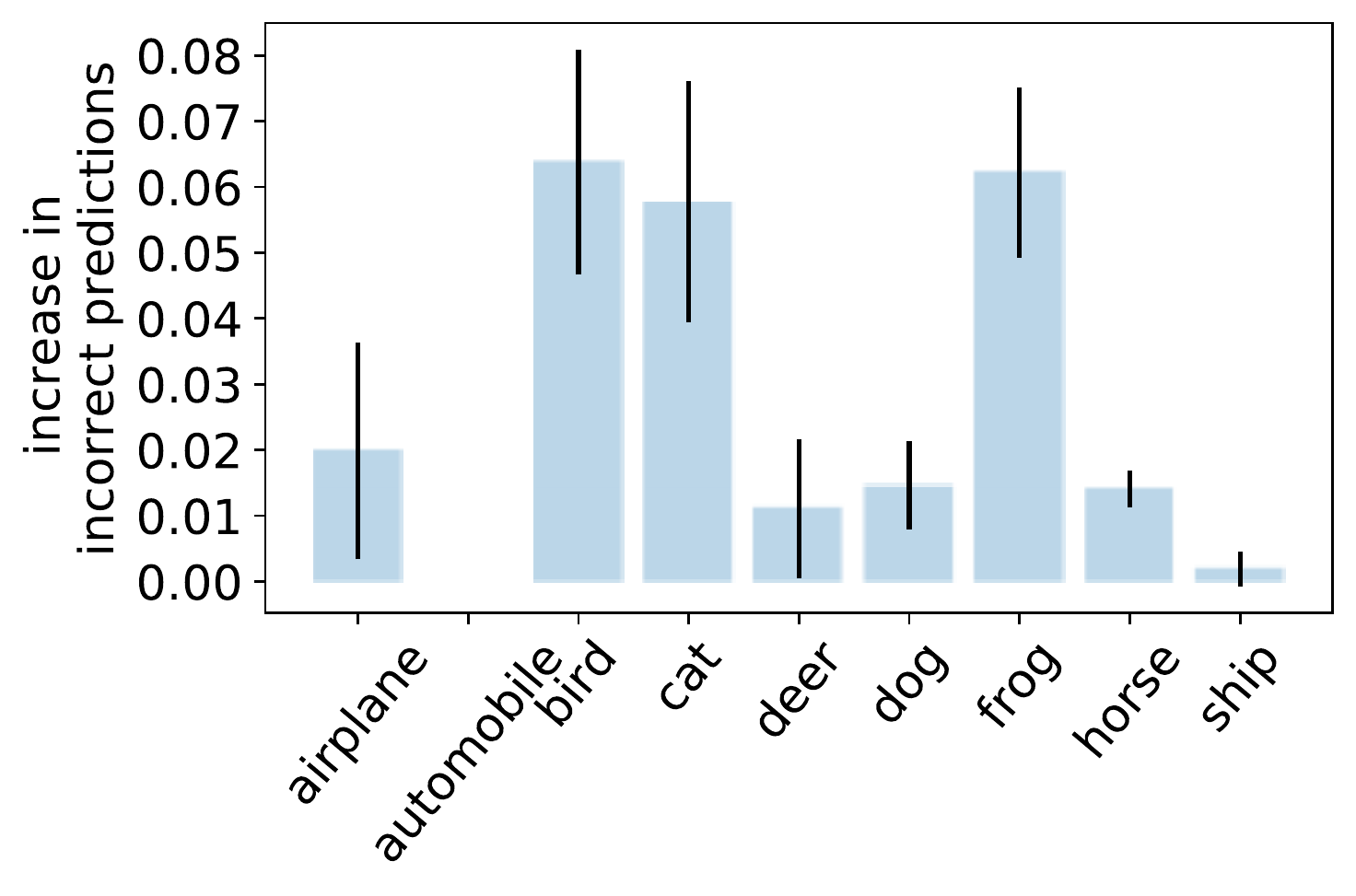}
    \end{minipage}
    \caption{Increase in incorrect in incorrect predictions for the basic (left) and \cutmix(right) models when removing the dominant class (``truck'') and occluding with non-uniform patches.}
    \label{fig:dominant}
\end{figure*}

We remove the 10th class from the \cifar{10} data set and retrain on the remaining classes.
We then evaluate the basic and \mixup models on data distorted with non-uniform patches. 
Just as in the other experiments, we obtain non-uniform patches by randomly sampling rectangular regions from the \cifar{100} images.
We once again find that the basic model show a bias towards one of the classes, namely the ``ship'' class (see Figure~\ref{fig:dominant}).
The basic model has a DI index of $1.11_{\pm 0.28}$, while \cutmix $0.16_{\pm0.04}$. 
Thus, when measuring robustness using \cutocc, the basic model will be disadvantaged.

\subsection{Results Obtained on a Data Set with Small Object Size}

All the benchmark data sets we evaluate robustness on have large, centred subjects. To provide a more comprehensive evaluation, we included a data set where the subject differs in scale and location for a classification task.
Since we were unable to find any publicly available data sets with these characteristics, we adapted \citet{DVN/7CBGOS_2019}'s data set to a binary classification task to identify the presence or absence of hard hats in the image samples. It must be noted that in the case of binary classification the bias is less clearly identifiable. In the multiclass case, we are looking for a consistent association  with a specific class across runs and distortion intensities. However in the binary case, any misclassification would naturally be assigned to the opposite class.

We evaluate the \cutocc and \iocc robustness levels for a level of occlusion of 30\% and present results in~\cref{tab:hat-occ}. 
The \cutocc robustness is massively overlapping , while \iocc reflects some of the added robustness of \fmix and \cutmix. 

\begin{table}[h]
    \centering
    \caption{Robustness to occlusion as measured using \cutocc and \iocc for the hard hat data set when occluding 30\% of the pixels in each image.} \label{tab:hat-occ}
    \vskip 0.15in
    \begin{tabulary}{\linewidth}{lLLLL}
    \toprule
         & basic & \fmix   & \cutmix  \\
    \midrule
    \cutocc & $67.95_{\pm 10.07}$ & $71.02_{\pm 8.49}$ & $69.59_{\pm 11.95}$ \\
    \iocc & $0.39_{\pm 0.37}$ & $1.47_{\pm 0.76}$ & $1.42_{\pm 0.51}$ \\
    \bottomrule
    \end{tabulary}
    \vskip -0.1in
\end{table}

\end{document}